\documentclass[journal]{IEEEtran}
\usepackage{times}
\usepackage{epsfig}
\usepackage{graphicx}
\usepackage{amsmath}
\usepackage{amssymb}
\DeclareUnicodeCharacter{FF0C}{fi}
\usepackage{multirow}
\usepackage{caption}
\usepackage{color}
\usepackage{float}
\usepackage{graphicx}
\usepackage{subfigure}
\usepackage[normalem]{ulem}
\begin{document}

\title{Generalizable Neural Radiance Fields for Novel View Synthesis with Transformer}

\author{Dan~Wang, Xinrui~Cui, Septimiu~Salcudean,~\IEEEmembership{Fellow,~IEEE} and Z.~Jane~Wang,~\IEEEmembership{Fellow,~IEEE}
%
%
}
\maketitle

\begin{abstract}
We propose a Transformer-based NeRF (TransNeRF) to learn a generic neural radiance field conditioned on observed-view images for the novel view synthesis task.
By contrast, existing MLP-based NeRFs are not able to directly receive observed views with an arbitrary number and require an auxiliary pooling-based operation to fuse source-view information, resulting in the missing of complicated relationships between source views and the target rendering view.
Furthermore, current approaches process each 3D point individually and ignore the local consistency of a radiance field scene representation.
These limitations potentially can reduce their performance in challenging real-world applications where large differences between source views and a novel rendering view may exist.
To address these challenges, our TransNeRF utilizes the attention mechanism to naturally decode deep associations of an arbitrary number of source views into a coordinate-based scene representation.
Local consistency of shape and appearance are considered in the ray-cast space and the surrounding-view space within a unified Transformer network.
Experiments demonstrate that our TransNeRF, trained on a wide variety of scenes, can achieve better performance in comparison to state-of-the-art image-based neural rendering methods in both scene-agnostic and per-scene finetuning scenarios especially when there is a considerable gap between source views and a rendering view.
\end{abstract}

\section{Introduction}
\label{sec:intro}
Novel view synthesis is a long-standing open problem concerned with the rendering of unseen views of a 3D scene given a set of observed views \cite{mildenhall2020nerf,wang2021ibrnet,8105827,6690212,9382843}.
It is a fundamental and challenging problem in 3D modeling \cite{9222285,7546862} and computer animation \cite{lingjie}.
The essence of novel view synthesis is to explore and learn a view-consistent 3D scene representation from a sparse set of input views.
The early work focused on modeling 3D shapes by discrete geometric 3D representations, such as mesh surface \cite{gkioxari2019mesh,nash2020polygen} point cloud \cite{qi2017pointnet} and voxel grid \cite{yang2018dense,Wang_2021_ICCV}.
Although explicit 3D geometry-based representations are intuitive, these representations are discrete and sparse. Therefore, they are incapable of learning high-resolution renderings with sufficient quality for complex real-world scenes.

\begin{figure}
	\centering
	\includegraphics[width=1\linewidth]{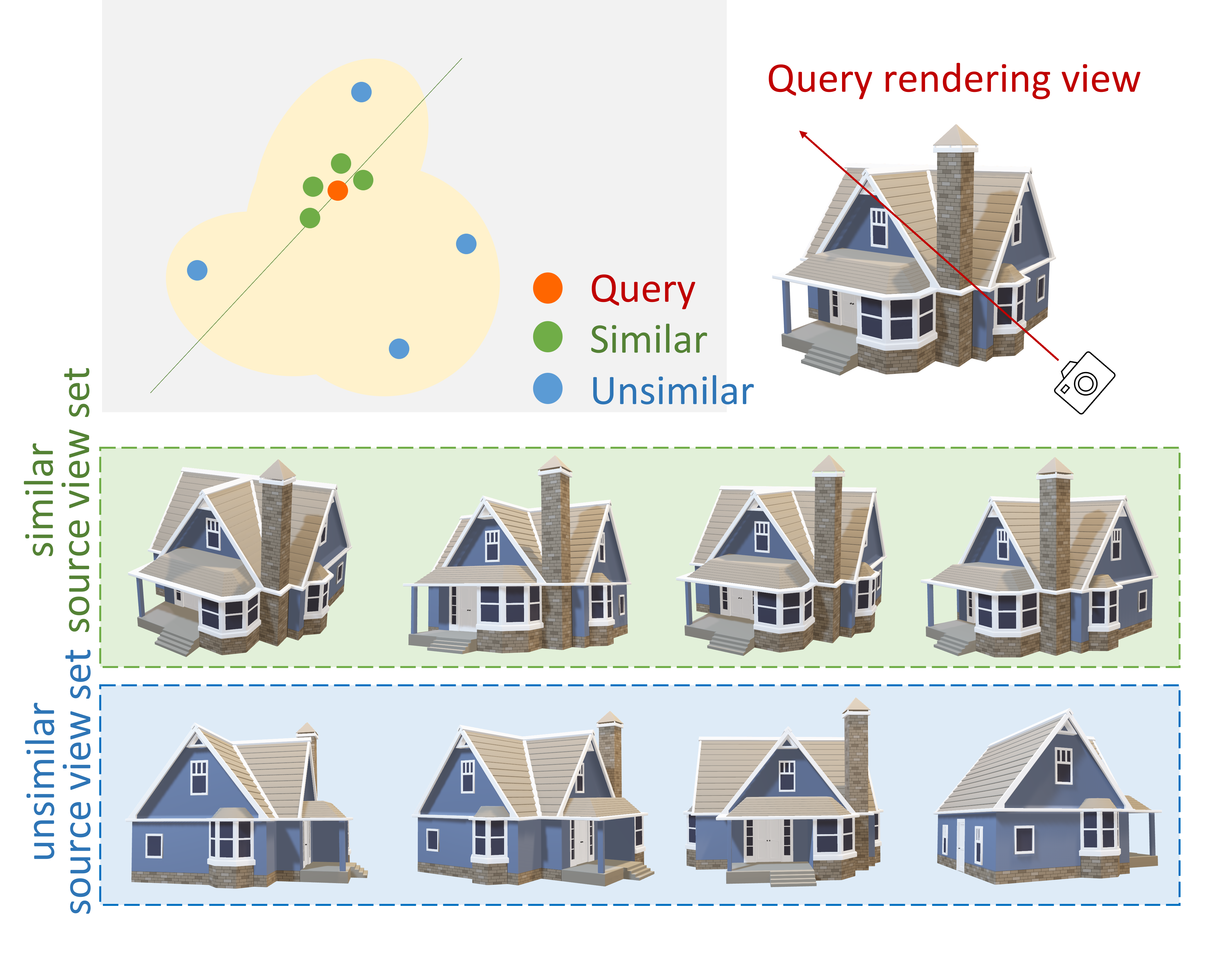}
	\centering
\caption{\textbf{Motivation}. 
{The complex radiance field scene representation function is shown with yellow/gray background. Given camera pose (rotation and translation) with respect to a world coordinate system, we can calculate the difference between each source view (the green and blue dots) and the target rendering view (the red one). The green ones are source views with small differences from the target one, which are distributed close to the target one in a 3D scene representation space, making it possible to approximate their relation by a linear function.
However, for source views (blue ones) with less similarity to the target view, their association is more complex and cannot be approximated well by a linear model.}
}

	\label{fig:TransNeRF1}
\end{figure}

The recent promising work, implicit continuous 3D coordinate-based representation \cite{mescheder2019occupancy,genova2020local}, \cite{park2019deepsdf,jiang2020local}, has been shown to have the potential to reconstruct 3D scenes with complicated high-resolution geometry and appearance. 
However, most discrete \cite{gkioxari2019mesh,nash2020polygen}, \cite{qi2017pointnet}, \cite{yang2018dense,Wang_2021_ICCV} and continuous \cite{mescheder2019occupancy,genova2020local}, \cite{park2019deepsdf,jiang2020local} 3D representations require ground-truth 3D geometry information as supervision, which is difficult to achieve in in real-world scenes.
Subsequently, research work \cite{insafutdinov2018unsupervised}, \cite{tulsiani2017multi}, \cite{sitzmann2020scene,niemeyer2020differentiable} re-visits this requirement by introducing differential approximation and rendering functions and optimizing the 3D representation only by multi-view images. 
But these approaches do not exploit the power of continuous 3D coordinate-based representation sufficiently and therefore only produce over-smoothed renderings.%

\begin{figure*}[]
	\centering
	\includegraphics[width=1\textwidth]{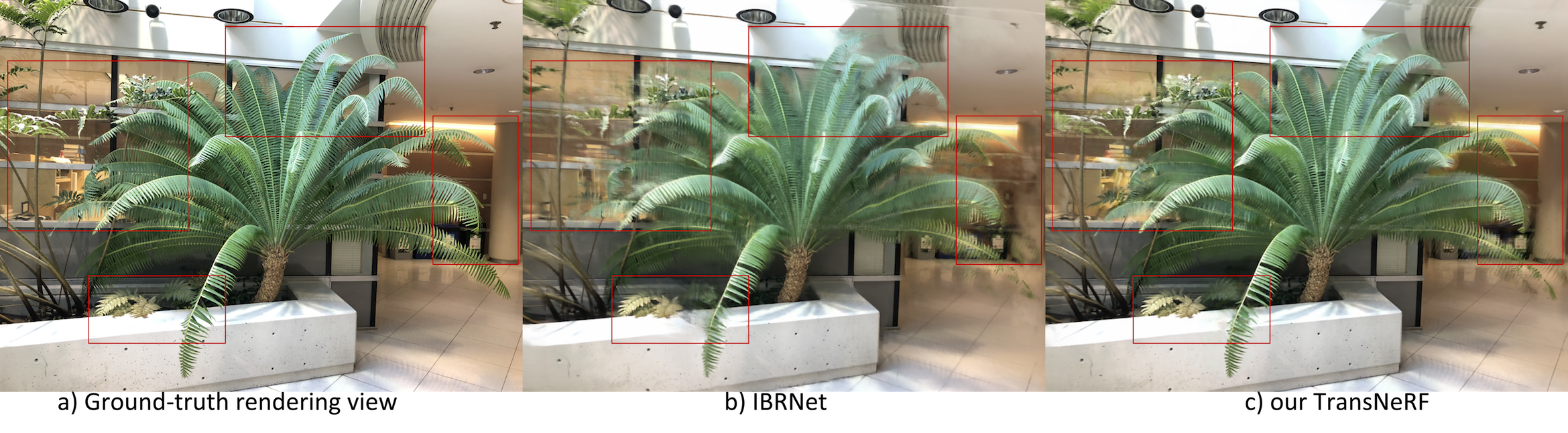}
	
\caption{\textbf{Example results from IBRNet \cite{wang2021ibrnet} (b) \textit{vs.} the proposed TransNeRF (c)} and ground-truth rendering view (a). We train a single scene-agnostic TransNeRF model on a large hybrid multi-scene dataset. It can effectively generalize to novel scenes without per-scene finetuning. In the same experimental setting for the scene-agnostic scenario, the result of our TransNeRF (c) is more realistic, containing fewer artifacts when compared with IBRNet (b).}
	\label{fig:TransNeRF2}
\end{figure*}

The more recent impressive Neural Radiance Field (NeRF)-based models \cite{mildenhall2020nerf}, as a sub-class of implicit continuous representation, introduce neural radiance field scene representations, which map a continuous 3D location and 2D viewing direction to its volume density and radiance color. 
Unlike previous models in implicit continuous representation, NeRF-based models can achieve high quality of novel view synthesis from observed images in realistic complex scenes. 
However, these models need to optimize a scene-specific 3D representation for each scene, which is time-consuming and does not learn the shared information amongst scenes.
Subsequently, to learn prior knowledge in diverse scenes, researchers \cite{yu2021pixelnerf,chen2021mvsnerf,wang2021ibrnet,trevithick2021grf} generalize the radiance field scene representation by receiving a pooling-based multi-view feature as a conditional input.

While these image-conditioned NeRFs \cite{yu2021pixelnerf,chen2021mvsnerf,wang2021ibrnet,trevithick2021grf} can generalize across different scenes, previous studies of NeRFs have not investigated in depth the relationship between observed source views and the target rendering view which is essential for novel view synthesis.
The reasons for this is that previous NeRF approaches were built on multi-layer perceptrons networks (MLPs) which are incapable of receiving and processing an arbitrary number of observed views; consequently, they use an auxiliary pooling-based model to aggregate multi-view features. This ignores in-depth and high-level complicated relationships across observed views and the rendering view.

This limitation impairs the potential of NeRFs to explore and learn a view-consistent 3D scene representation from observed views, especially in practical applications where the observed source views might be captured at camera poses that are very different from the camera pose of the target view.
As illustrated in Fig \ref{fig:TransNeRF1}, when camera poses of source views are similar to the rendering view, source views and the target view are distributed in a local region in 3D scene representation space, making it possible to approximate their relationship by a linear function as in previous techniques \cite{yu2021pixelnerf,chen2021mvsnerf,wang2021ibrnet,trevithick2021grf}.
However, as the difference between observed views and the rendered view increases, the correlation becomes more complicated, making it challenging for these approaches to synthesize a novel view.
Therefore, in this scenario, prior MLP-based NeRFs \cite{yu2021pixelnerf,chen2021mvsnerf,wang2021ibrnet,trevithick2021grf}, using the pooling-based function to fuse the multi-view, fail to resolve this challenge.

\begin{figure*}
	\centering
	\includegraphics[width=1\linewidth]{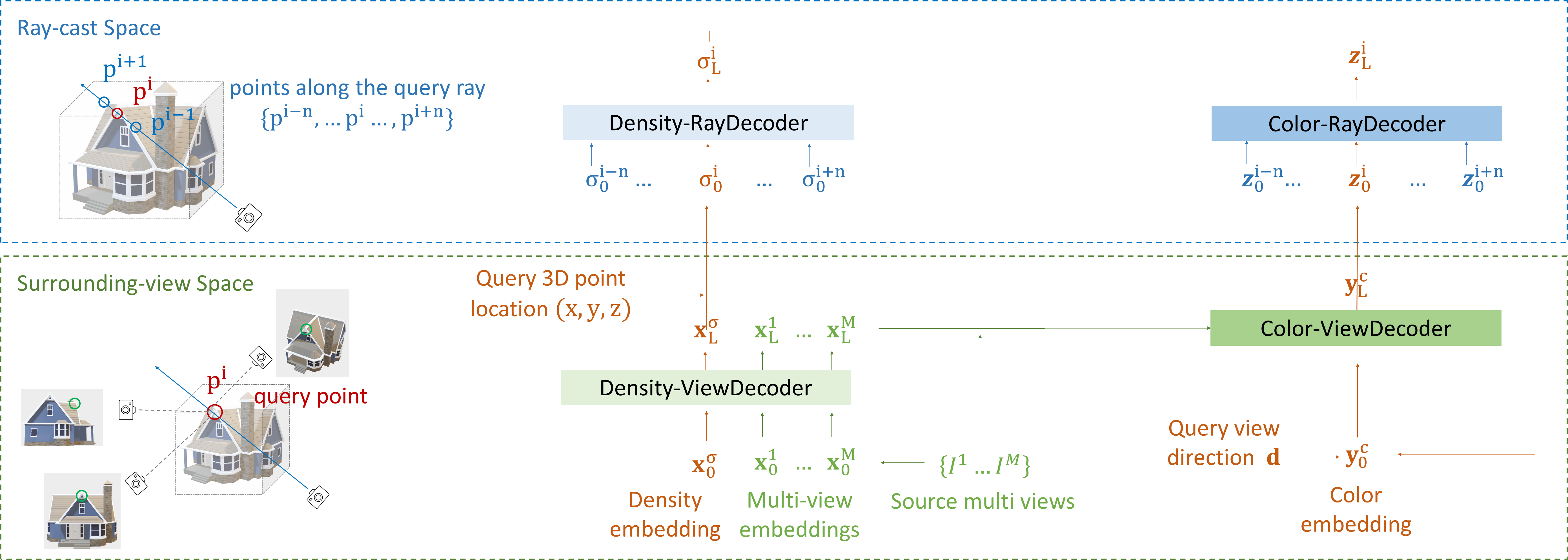}
	\centering
\caption{For rendering a query 3D point on a target-viewing ray, the proposed TransNeRF consists: 1) In the surrounding-view space, our Density-ViewDecoder(Sec. \ref{Density-ViewDecoder}) and Color-ViewDecoder(Sec. \ref{Color-ViewDecoder}) fuse source views and query spatial information ($(x,y,z)$, $\mathbf{d}$) into the latent density and color representations for the query point;
2) In the ray-cast space, we use Density-RayDecoder(Sec. \ref{Density-RayDecoder}) and Color-RayDecoder(Sec. \ref{Color-RayDecoder}) to enhance the query density and color representations by considering neighboring points along the target view ray.
Finally, we can obtain the volume density and directional color for the query 3D point on a target-viewing ray from our TransNeRF.}
	\label{fig:TransNeRF-flow}
\end{figure*}

To tackle this unmet need, in this paper, we propose a Transformer-based NeRF framework (TransNeRF) utilizing its powerful attention mechanism to learn a general continuous 3D scene representation from an arbitrary number of observed views.
The learning process of TransNeRF is divided into two stages: the first deals with the surrounding-view space and the second deals with in the ray-cast space, shown in Fig \ref{fig:TransNeRF-flow}.
In the surrounding-view space, unlike previous MLP-based work requiring an auxiliary pooling-based operation to fuse source views, our framework is only built on the Transformer-based network and leverages its attention mechanism to integrate deep relationships between the rendering view and observed views as a coordinate-based scene representation. Specifically, in our attention layers at our Density-ViewDecoder and Color-ViewDecoder, TransNeRF can naturally decode observed views and spatial location information for a query 3D point in the rendering ray direction into its density and directional color representations.
In the ray-cast space, when rendering a query 3D-point, our TransNeRF considers adjacent 3D points on the target ray simultaneously by attention layers in Density-RayDecoder and Color-RayDecoder.
In contrast, in prior work each 3D point is processed independently. 
In this paper, taking advantage of the attention mechanism, our proposed approach enhances the local consistency of a 3D scene representation in both the ray-cast space and the surrounding-view space by a unified Transformer-based network.
Therefore, the proposed TransNeRF has the capability to learn a more comprehensive neural radiance field that can effectively generalize to novel scenes without per-scene finetuning, as shown in Fig. \ref{fig:TransNeRF2}.

Our contributions can be summarized as follows:
\begin{itemize}
\item We introduce TransNeRF, a unified Transformer-based architecture, to model a general neural radiance field from observed views for novel view synthesis, whereas previous MLP-based NeRFs need an auxiliary pooling model to aggregate multi-view information when dealing with an arbitrary number of observed source views.
\item TransNeRF, utilizing the attention mechanism, integrates information of the projected 2D pixels in the surrounding source views and the neighboring 3D points along the query ray when rendering the density and radiance color of a query 3D point. It gives our model a better understanding of the shape and appearance consistency in a local neural radiance field.
\item TransNeRF can explore complicated relationships between observed views and the rendering view and fuses the learned high-level multi-view information into coordinate-based 3D scene representation by our attention layers.
\item Experiments demonstrate that in both scene-agnostic and per-scene finetuning experimental settings, TransNeRF achieves more realistic rendering results than previous state-of-the-art methods, especially when source views are captured at camera poses with a considerable difference to the camera pose of the rendering view.

\end{itemize}
In this paper, we present a generic Transformer-based NeRF framework with a high capacity in modeling radiance field scene representation from observed source view images.
Similar to MLP-based NeRFs, our framework can be easily extended to advanced TransNeRF derivatives, e.g. from NeRF to NeRF-W \cite{martin2021nerf} by adding appearance and transient variables as inputs to model uncertainty in the wild scenarios.
Here, we only focus on the vanilla TransNeRF and hope this work will serve as a blueprint for future work.

\section{related work}

\subsection{Novel View Synthesis}
The goal of novel view synthesis is to reconstruct unseen views of a scene from its observed multiple images.
For this task, solutions can be separated into three categories: novel view rendering by image-based representation, discrete 3D shape-based representation, and implicit continuous 3D spatial coordinate-based representation.

\textbf{Image-based representation.}
The image-representation-based approaches leverage GAN models and auto-encoders  \cite{goodfellow2014generative} to learn a global embedding from multi-view images and render an unseen view given its camera parameters. 
However, these image-representation-based models are limited by the lack of understanding of the underlying 3D shape representation.
In contrast, the two 3D-representation-based types reconstruct novel views by exploiting multi-view consistent 3D shape representation.

\textbf{Discrete 3D shape-based representation.}
For novel view synthesis, the discrete 3D-shape representation involves point cloud \cite{qi2017pointnet}, \cite{insafutdinov2018unsupervised}, mesh surface \cite{gkioxari2019mesh,nash2020polygen}, voxel grid \cite{yang2018dense,Wang_2021_ICCV}, \cite{tulsiani2017multi} approaches.
The point-cloud-based representation \cite{qi2017pointnet} reconstructed from multi-view images is typically sparse, making it difficult to infer the missing information.
The mesh-based representation \cite{gkioxari2019mesh,nash2020polygen} requires a template mesh with fixed topology and rigid texture parameterizations for the category-specific shape. In complex real-world scenes, it is challenging to fulfill this requirement.
The volume-based representation \cite{yang2018dense,Wang_2021_ICCV} can easily leverage CNNs to obtain predicted voxel grids. 
Nevertheless, it causes discretization artifacts from low-resolution voxel grids and limits its resolution-scale ability with computational and memory constraints.

\textbf{Implicit continuous 3D spatial coordinate-based representation.}
Recently, another promising branch implicitly represents geometric and appearance information of a 3D point in a scene as a function of its continuous spatial coordinates. 
Therefore this implicit continuous 3D shape representation has the potential to render higher-resolution complicated geometry and appearance of scenes without storing an entire scene representation.
In this direction, research work modeling occupancy fields \cite{mescheder2019occupancy,genova2020local} or signed distance functions \cite{park2019deepsdf,jiang2020local} requires ground-truth 3D information for supervision, limiting its real-world applications. 
To relax this requirement, researchers \cite{sitzmann2020scene,niemeyer2020differentiable} leverage differential rendering functions to allow 2D images as only supervision information for 3D representation. 
However, these models are unable to fully exploit the advantage of the implicit continuous representation, which limits it to simple geometric and appearance scenarios.

\textbf{Neural radiance field scene representation.}
More recently, the impressive neural radiance fields (NeRF) \cite{mildenhall2020nerf} have shown a solid ability to synthesize novel views by representing continuous scenes as 5D radiance fields in MLPs.
Nevertheless, NeRF optimizes each scene representation independently, not exploring the shared information amongst scenes and being time-consuming.
To solve this, researchers proposed models, such as PixelNeRF \cite{yu2021pixelnerf}, MVSNeRF \cite{chen2021mvsnerf}, IBRNet \cite{wang2021ibrnet}, which receives as conditional inputs multiple observer views to learn a general neural radiance field.
Followed the design principle of divide-and-conquer, they have two separate components: a CNN feature extractor for a single image and a MLP as a NeRF network.
For a single-view stereo, in these models, CNNs map an image to a feature grid, and MLPs map a query 5D coordinate and its corresponding CNN feature to a single volume density and view-dependent RGB color. 
For a multi-view stereo, since CNN and MLP are unable to process an arbitrary number of input views, they first process the coordinates and corresponding features in each view coordinate frame independently and obtain an image-conditioned intermediate representation for each view. 
Next, they use an auxiliary pooling-based model to aggregate these intermediate representations across the views within the NeRF network.
In the 3D understanding, multiple views provide additional information about the scene.

Nonetheless, pooling-based fusion models in these methods can barely explore the complex relationship across multiple views for 3D scene understanding.
Furthermore, processing each 3D point independently ignores the local consistency of a 5D radiance field in a scene.
To tackle it, we propose an encoder-decoder Transformer framework, TransNeRF, to represent the neural radiance field scene. 
Compared with the pooling-based multi-view representation in previous work, TransNeRF can explore deep relationships amongst multiple views and aggregate multi-view information into the coordinate-based scene representation by the attention mechanism in a single Transformer-based NeRF.
Furthermore, TransNeRF can learn the local geometry consistency of shape and appearance in a scene by considering the corresponding information in the ray-cast space and the surrounding-view space.

\subsection{Transformer}
Transformer recently emerged as a promising network framework and achieved impressive performance in natural language processing \cite{brown2020language} and computer vision \cite{carion2020end, dosovitskiy2020image, chen2020generative}. 
The main idea behind this approach is to utilize the multi-head self-attention operation to explore the dependence within input tokens and then learn a global feature representation. 
In the object detection task, DETR \cite{carion2020end} presents a new framework combining a 2D CNN with a Transformer and predicts object detection in parallel as a sequence of output tokens. 
In image classification, ViT \cite{dosovitskiy2020image} demonstrates the impressive ability to learn global contexts in Transformer even without using CNN features that are more focused on local concepts \cite{9491931}.

For novel view synthesis, we introduce an end-to-end Transformer framework to implicitly model the continuous 3D scene as a neural radiance field representation. Our model leverages the Transformer advantage of exploring deep relationships among observed images to learn a view-consistent 3D representation.

\section{Methodology}

Unlike Neural Radiance Field (NeRF) \cite{mildenhall2020nerf} optimizing a scene-specific 3D representation, we propose TransNeRF to learn a generic radiance field representation for novel scenes.
Given captured multi-view images $\lbrace\mathbf{I}^m\rbrace^M_{m=1}$ ($M$ source views) of diverse scenes and their camera parameters $\lbrace \mathbf{\Theta}^m\rbrace^M_{m=1}$ (camera poses, intrinsic parameters and scene bounds), TransNeRF reconstructs a generic radiance field, denoted $F_{\text{TransNeRF}}$, learning prior knowledge shared across scenes
\begin{align}
(\sigma,\mathbf{c})\leftarrow F_{\text{TransNeRF}}((x,y,z),\mathbf{d};\lbrace\mathbf{I}^m, \mathbf{\Theta}^m\rbrace_m) \,\,,
\end{align}
where $(x,y,z)$ is a 3D point location, $\mathbf{d}$ denotes a unit-length viewing ray-direction and outputs are a differential volumetric density $\sigma$ and a directional emitted color $\mathbf{c}$.

To render a query 3D point on a target-view ray, TransNeRF receives its spatial location $(x,y,z)$, viewing direction $\mathbf{d}$ and a set of source views $\lbrace\mathbf{I}^m, \mathbf{\Theta}^m\rbrace_m$ as inputs.
The TransNeRF framework contains the following four steps:
1) For each observed source view, we first extract its feature volume from a pre-trained U-Net and retrieve spatial-aligned 2D-pixel features for the query 3D point as its initial source-view feature. 
Then, \textbf{Density-ViewDecoder} (Sec. \ref{Density-ViewDecoder}) decodes the initial features for source views into the latent density representations of the query point.
2) \textbf{Density-RayDecoder} (Sec. \ref{Density-RayDecoder}) thereafter enhances the latent density representation by considering neighboring points along the target-view ray and outputs the density for the query 3D point.
3) Subsequently, given latent source-view features from Density-ViewDecoder and the query-point density representation from Density-RayDecoder, \textbf{Color-ViewDecoder} (Sec. \ref{Color-ViewDecoder}) fuses them into the latent color representation for the query point.
4) Similar to Density-RayDecoder, \textbf{Color-RayDecoder} (Sec. \ref{Color-RayDecoder}) decodes the query color by integrating color information of 3D points along the target-viewing ray.
Finally, we can obtain the density and directional color $(\sigma,\mathbf{c})$ for the query 3D point on a target-viewing ray.

As in NeRF \cite{mildenhall2020nerf}, classical volume rendering \cite{kajiya1984ray}, a differentiable ray marching rendering, is then utilized to render a projection 2D image from our radiance field scene representation $F_{\text{TransNeRF}}$.
To calculate a 2D-pixel’s color at a rendered image, classical volume rendering marches a ray via the pixel and accumulates radiance at sampled 3D points along the ray.
Specifically, the rendered color $C(\mathbf{r})$ of the camera ray $\mathbf{r}(t)=\mathbf{o}+t\mathbf{d}$ with near and far bounds $[{t_n}, {t_f}]$ is computed as:
\begin{align}\label{Eq-VR}
&C(\mathbf{r})=\int_{t_n}^{t_f}T(t)\sigma(\mathbf{r}(t);\mathbf{I}^m, \mathbf{\Theta}^m)\mathbf{c}(\mathbf{r}(t),\mathbf{d};\mathbf{I}^m, \mathbf{\Theta}^m)dt 
\end{align}
where the function $T(t)=\exp(-\int_{t_n}^{t}\sigma(\mathbf{r}(s);\mathbf{I}^m, \mathbf{\Theta}^m)ds)$ calculates the accumulated transmittance along the ray between depth bounds $[{t_n}, {t_f}]$.

The rendering loss between the rendered and ground truth pixel colors for all the camera rays of the target view with camera pose $\mathbf{\Theta}$ is
\begin{align}\label{render}
\mathcal{L}=\sum_{\mathbf{r}\in\mathcal{R}(\mathbf{\Theta})}\lVert\hat{C}(\mathbf{r}) - C(\mathbf{r})\rVert_2^2   \,\,,
\end{align}
where $\mathcal{R}(\mathbf{\Theta})$ is the set of all camera rays of the desired virtual camera with pose $\mathbf{\Theta}$.

Given a set of observed source views, the rendering loss $\mathcal{L}$ between the observations and the predictions is minimized by optimizing the parameters of the neural radiance field $F_{\text{TransNeRF}}$.
Here, our TransNeRF is fully differentiable and can be trained end-to-end only requiring source views. After being trained on a large dataset containing diverse scenes, our TransNeRF effectively generalizes to unseen scenes even without per-scene finetuning.
Accordingly, given the camera pose of a query novel view of a scene and its multiple source views, our TransNeRF can render the query novel view from the pre-trained general neural radiance field by casting rays from the target camera center towards the 3D space using volume rendering in Eq. \ref{Eq-VR}.

\subsection{Density Decoder in Surrounding-view Space}\label{Density-ViewDecoder}

We first present our density decoder in surrounding-view space (Density-ViewDecoder) fusing source views into the latent volume-density representation for a query 3D point.

For each source view image, we first extract its feature volume by a pre-trained view-shared U-Net.
A query 3D point $(x,y,z)$ is then projected into each source-view image $\mathbf{I}^m$ by its camera projection matrix $\mathbf{\Theta}^m$ to extract the corresponding RGB color $\lbrace\mathbf{c}_{src}^m\rbrace^M_{m=1}$ and feature vector $\lbrace\mathbf{f}_{src}^m\rbrace^M_{m=1}$ at the projected 2D pixel location through bilinear interpolation.
In each source view, we also record its viewing direction $\lbrace\mathbf{d}_{src}^m\rbrace^M_{m=1}$ for the projected 2D pixel from the source camera pose.
And then based on these information, we obtain the initial source-view embeddings $\lbrace\mathbf{x}_{0}^m\rbrace^M_{m=1}$ for source views, as in \cite{wang2021ibrnet}. 

For the query 3D point, Density-ViewDecoder receives the initial source-view embeddings $\lbrace\mathbf{x}_{0}^m\rbrace^M_{m=1}$ and the learnable query density embedding $\mathbf{x}^{\sigma}_0$ as inputs $\mathbf{X}_0$.
The Density-ViewDecoder can be formulated as follows: 
\begin{align}
& \mathbf{X}_0=[\mathbf{x}^{\sigma}_0;\mathbf{x}^{1}_0; \mathbf{x}^{2}_0; \cdots; \mathbf{x}^{M}_0]\\
&\mathbf{\tilde{X}}_{l}=\text{Norm}(\text{Density-ViAttn}(\mathbf{X}_{l-1})+\mathbf{X}_{l-1})\label{den}\\
&\mathbf{X}_{l}=\text{Norm}(\text{FFN}(\mathbf{\tilde{X}}_{l})+\mathbf{\tilde{X}}_{l}) \,\,,
\end{align}
where $l$ denotes the index of a basic block ($l=1,\cdots,L$), ``Norm'' is the layer normalization function and ``FFN'' is a position-wise feed-forward network. 
At the L-th block, we can obtain $\mathbf{X}_L=[\mathbf{x}^{\sigma}_L;\mathbf{x}^{1}_L; \mathbf{x}^{2}_L; \cdots; \mathbf{x}^{M}_L]$.
In the Density-ViewDecoder, we concatenate the embedding $\mathbf{x}_L^\sigma$ and its 3D coordinate location $(x,y,z)$ as the output density representation for the query 3D point.

Density-view attention layers (Density-ViAttn) can explore deep relationships among source views, defined as follows:
\begin{align}\label{self}
\text{Density-ViAttn}(\mathbf{X})&=\text{MH-Attn}(\mathbf{X},\mathbf{X},\mathbf{X}) \,\,,
\end{align}
where the multi-head attention function is defined as:
\begin{align} \label{MH}
&\text{MH-Attn}(\mathbf{Q},\mathbf{K},\mathbf{V})=\text{Cat}(\mathbf{A}_{1},\cdots,\mathbf{A}_{H})\mathbf{W} \,\,,\\
&\text{where }
\mathbf{A}_{h}=\text{Attention}(\mathbf{Q}_h,\mathbf{K}_h,\mathbf{V}_h) \nonumber\\
&\quad\quad\quad\mathbf{Q}_h=\mathbf{Q}\mathbf{W}^q_h;\mathbf{K}_h=\mathbf{K}\mathbf{W}_h^k;\mathbf{V}_h=\mathbf{V}\mathbf{W}^v_h \,\,. \nonumber
\nonumber
\end{align}
Here, $N_q$ queries are stacked in $\mathbf{Q}\in\mathbb{R}^{N_q\times d_k}$, a set of $N_{kv}$ key-value pairs are stacked in $\mathbf{K}\in\mathbb{R}^{N_{kv}\times d_{k}}$ and $\mathbf{V}\in\mathbb{R}^{N_{kv}\times d_{v}}$.
And $\mathbf{W}^q_h,\mathbf{W}^k_h\in\mathbb{R}^{d_{k}\times d_{h}}; \mathbf{W}^v_h\in\mathbb{R}^{d_{v}\times d_{h}}$ and $\mathbf{W}\in\mathbb{R}^{Hd_h\times d_k}$ are parameter matrices ($H\times d_h=d_k$ and $d_h$ is the feature dimension in each head).
And the Attention function is computed by
\begin{equation}
\begin{split}
&\text{Attention}(\mathbf{Q}_h, \mathbf{K}_h, \mathbf{V}_h) = \text{softmax}(\frac{\mathbf{Q}_h\mathbf{K}_h^T}{\sqrt{d_{k}}}) \mathbf{V}_h \,\,,
\end{split}
\end{equation}
where attention-score, stored in $\text{softmax}(\frac{\mathbf{Q}_h\mathbf{K}_h^T}{\sqrt{d_{k}}})$, of a specific value is obtained by the match between this query and the key paired with the target value.
Our Density-ViewDecoder is invariant to permutations of source views and can receive an arbitrary number of source views.

\subsection{Density Decoder in Ray-cast Space}\label{Density-RayDecoder}

The density decoder in ray-cast space (Density-RayDecoder) decodes density information of the query 3D point by aggregating the density features of the neighboring 3D points along the target-view ray.

For the query point $q^{i}$ and neighboring $2n$ points $\lbrace q^{i-n}, \dotsm, q^{i-1}, q^{i+1}, \dotsm, q^{i+n}\rbrace$ along the target-viewing ray, we denote $[\mathbf{\sigma}_0^{i-n};\dotsm;\mathbf{\sigma}_0^{i-1};\mathbf{\sigma}_0^{i};\mathbf{\sigma}_0^{i+1};\dotsm;\mathbf{\sigma}_0^{i+n}]$ as their initial density representations at the input end of the Density-RayDecoder.
Here, the initial density representation for each 3D point is computed by a linear function of the Density-ViewDecoder output for the corresponding point ($\mathbf{\sigma}_0=\text{FC}(\mathbf{x}_L^\sigma \odot(x,y,z))$, where $\odot$ is the concatenation operation).
And then positional encodings $\mathbf{E}^{pos}$ are added to density representations of neighboring 3D points to keep their position information in the ray-cast space. 
Each positional encoding informs each point of its 3D spatial location, which is computed by utilizing sine and cosine functions of different frequencies in a similar way to \cite{carion2020end}.
The Density-RayDecoder is represented as: 
\begin{align}
&\mathbf{D}_0=[\mathbf{\sigma}_0^{i-n};\dotsm;\mathbf{\sigma}_0^{i-1};\mathbf{\sigma}_0^{i};\mathbf{\sigma}_0^{i+1};\dotsm;\mathbf{\sigma}_0^{i+n}]+\mathbf{E}^{pos}\\
&\mathbf{\tilde{D}}_{l}=\text{Norm}(\text{Density-Attn}(\mathbf{D}_{l-1})+\mathbf{D}_{l-1})\\
&\mathbf{D}_{l}=\text{Norm}(\text{FFN}(\mathbf{\tilde{D}}_{l})+\mathbf{\tilde{D}}_{l}) \,\,,
\end{align}
where the density attention layer (Density-Attn) is computed as $\text{Density-Attn}(\mathbf{D})=\text{MH-Attn}(\mathbf{D},\mathbf{D},\mathbf{D})$ fusing information of surrounding 3D points on the target-viewing ray.
Finally, at the end block, the Density-RayDecoder outputs the density representation $\mathbf{\sigma}_L^{i}$ of the query 3D point and then we use a linear function to project it to the density value for the query point.

\subsection{Color Decoder in Surrounding-view Space}\label{Color-ViewDecoder}

The color decoder in surrounding-view space (Color-ViewDecoder) learns a query directional emitted color as a function of viewing direction, source-view representations learned from Density-ViewDecoder and the latent density representation from Density-RayDecoder.

The Color-ViewDecoder can be formulated as follows: 
\begin{align}
&\mathbf{\tilde{Y}}_{l}=\text{Norm}(\text{Color-ViAttn}(\mathbf{Y}_{l-1}, \mathbf{X}_{src},\mathbf{C}_{src})+\mathbf{Y}_{l-1}) \label{VC}\\
&\mathbf{Y}_{l}=\text{Norm}(\text{FFN}(\mathbf{\tilde{Y}}_{l})+\mathbf{\tilde{Y}}_{l}) \,\,.
\end{align}
In the color-view layer (Color-ViAttn), the initial query directional color embedding is represented as $\mathbf{Y}_0=\text{FC}(\mathbf{\sigma}_L^{i})\odot\mathbf{d}_{tgt}$, where $\mathbf{\sigma}_L^{i}$ is the latent density representation from Density-RayDecoder and $\mathbf{d}_{tgt}$ is the target-viewing direction for the query 3D point. The Color-ViAttn layer is calculated as:
\begin{align}
\text{Color-ViAttn}(\mathbf{Y}, \mathbf{X}_{src},\mathbf{C}_{src})=\text{MH-Attn}(\mathbf{Y}, \mathbf{X}_{src},\mathbf{C}_{src}) \,\,,
\end{align}
where the value is $\mathbf{C}_{src}=[\gamma(\mathbf{c}^{1}_{src}); \gamma(\mathbf{c}^{2}_{src}); \cdots; \gamma(\mathbf{c}^{M}_{src})]$  ($\gamma(\cdot)$ is the embedding function) and the key is $\mathbf{X}_{src}=[\text{FC}(\mathbf{x}^{1}_L)\odot\mathbf{d}^1_{src}; \text{FC}(\mathbf{x}^{2}_L)\odot\mathbf{d}^2_{src}; \cdots; \text{FC}(\mathbf{x}^{M}_L)\odot\mathbf{d}^M_{src}]$ representing conditional source-view information.
The output of Color-ViewDecoder $\mathbf{y}_L^c$ is the latent color representation for the query 3D point.

\subsection{Color Decoder in Ray-cast Space}\label{Color-RayDecoder}

The color decoder in ray-cast space (Color-RayDecoder) learns a query directional emitted color by fusing color features of adjacent 3D points along the target ray in color attention layers ($\text{Color-Attn}(\mathbf{Z})=\text{MH-Attn}(\mathbf{Z},\mathbf{Z},\mathbf{Z})$).
The Color-RayDecoder can be formulated as follows:
\begin{align}
&\mathbf{Z}_0=[\mathbf{z}_0^{i-n};\dotsm;\mathbf{z}_0^{i-1};\mathbf{z}_0^{i};\mathbf{z}_0^{i+1};\dotsm;\mathbf{z}_0^{i+n}]+\mathbf{E}^{pos}\\
&\mathbf{\tilde{Z}}_{l}=\text{Norm}(\text{Color-Attn}(\mathbf{Z}_{l-1})+\mathbf{Z}_{l-1})\\
&\mathbf{Z}_{l}=\text{Norm}(\text{FFN}(\mathbf{\tilde{Z}}_{l})+\mathbf{\tilde{Z}}_{l}) \,\,.
\end{align}
where the latent color representation $\mathbf{y}_L^c$ for the query 3D point is assigned to the corresponding $\mathbf{z}_0^{i}$ and likewise for latent color representations for adjacent $2n$ 3D-points in local ray-cast space.

Subsequently, after the Color-RayDecoder, we use a linear function to project the output color embedding $\mathbf{z}_L^{i}$ to its output predicted color value. 
Then the predicted volume density and directional color of a query point along a camera ray of the desired virtual camera are put forward to the classical volume rendering in Eq.(\ref{render}).
The implementation details of the network and training are described in the supplemental material.

\begin{table*}[]
\centering
\resizebox{\textwidth}{!}{
\begin{tabular}{cc|cccc|cccc|cccc} \hline
\multicolumn{14}{c}{Scene-agnostic setting on realistic synthetic dataset \cite{mildenhall2020nerf}}                                                                                                                        \\ \hline
\multicolumn{1}{l}{}       & \multicolumn{1}{l}{} & \multicolumn{4}{c}{PSNR $\uparrow$}              & \multicolumn{4}{c}{SSIM $\uparrow$}              & \multicolumn{4}{c}{LPIPS $\downarrow$}           \\ \hline
Scene                      & $\mathbf{S}_i$             & PixelNeRF & MVSNeRF & IBRNet & TransNeRF      & PixelNeRF & MVSNeRF & IBRNet & TransNeRF      & PixelNeRF & MVSNeRF & IBRNet & TransNeRF      \\ \hline
\multirow{4}{*}{Chair}     & S1                   & 21.21     & 23.50   & 28.55  & \textbf{29.06} & 0.890     & 0.910   & 0.942  & \textbf{0.954} & 0.135     & 0.108   & 0.066  & \textbf{0.055} \\
                           & S2                   & 16.98     & 19.38   & 24.93  & \textbf{25.79} & 0.784     & 0.811   & 0.854  & \textbf{0.903} & 0.240     & 0.224   & 0.173  & \textbf{0.135} \\
                           & S3                   & 15.76     & 18.37   & 24.12  & \textbf{25.17} & 0.712     & 0.745   & 0.798  & \textbf{0.870} & 0.298     & 0.267   & 0.206  & \textbf{0.164} \\
                           & S4                   & 14.51     & 17.21   & 23.12  & \textbf{24.85} & 0.612     & 0.647   & 0.714  & \textbf{0.823} & 0.387     & 0.338   & 0.259  & \textbf{0.194} \\ \hline
\multirow{4}{*}{Lego}      & S1                   & 19.51     & 21.13   & 24.64  & \textbf{26.13} & 0.834     & 0.862   & 0.917  & \textbf{0.923} & 0.174     & 0.148   & 0.106  & \textbf{0.090} \\
                           & S2                   & 15.30     & 17.08   & 21.14  & \textbf{22.80} & 0.685     & 0.717   & 0.787  & \textbf{0.851} & 0.331     & 0.311   & 0.257  & \textbf{0.179} \\
                           & S3                   & 14.21     & 16.08   & 20.33  & \textbf{22.18} & 0.613     & 0.650   & 0.731  & \textbf{0.819} & 0.390     & 0.354   & 0.291  & \textbf{0.207} \\
                           & S4                   & 13.04     & 14.92   & 19.33  & \textbf{21.10} & 0.532     & 0.571   & 0.665  & \textbf{0.768} & 0.471     & 0.421   & 0.339  & \textbf{0.246} \\ \hline
\multirow{4}{*}{Ship}      & S1                   & 21.31     & 21.79   & 22.92  & \textbf{24.57} & 0.803     & 0.808   & 0.825  & \textbf{0.836} & 0.267     & 0.256   & 0.227  & \textbf{0.211} \\
                           & S2                   & 17.04     & 17.69   & 19.37  & \textbf{21.25} & 0.650     & 0.658   & 0.689  & \textbf{0.725} & 0.443     & 0.438   & 0.397  & \textbf{0.305} \\
                           & S3                   & 15.98     & 16.69   & 18.56  & \textbf{20.64} & 0.577     & 0.591   & 0.633  & \textbf{0.693} & 0.503     & 0.480   & 0.431  & \textbf{0.333} \\
                           & S4                   & 14.83     & 15.59   & 17.63  & \textbf{19.34} & 0.481     & 0.497   & 0.552  & \textbf{0.637} & 0.599     & 0.565   & 0.495  & \textbf{0.370} \\ \hline
\multirow{4}{*}{Drums}     & S1                   & 17.60     & 18.74   & 21.28  & \textbf{21.99} & 0.771     & 0.807   & 0.893  & \textbf{0.902} & 0.184     & 0.164   & 0.113  & \textbf{0.105} \\
                           & S2                   & 13.85     & 15.14   & 18.21  & \textbf{18.69} & 0.642     & 0.683   & 0.782  & \textbf{0.802} & 0.321     & 0.312   & 0.252  & \textbf{0.201} \\
                           & S3                   & 12.72     & 14.16   & 17.40  & \textbf{18.49} & 0.570     & 0.616   & 0.725  & \textbf{0.770} & 0.383     & 0.354   & 0.285  & \textbf{0.230} \\
                           & S4                   & 11.53     & 12.99   & 16.40  & \textbf{18.18} & 0.480     & 0.528   & 0.652  & \textbf{0.724} & 0.474     & 0.434   & 0.346  & \textbf{0.254} \\ \hline
\multirow{4}{*}{Mic}       & S1                   & 26.51     & 27.28   & 28.93  & \textbf{29.32} & 0.929     & 0.938   & 0.951  & \textbf{0.968} & 0.071     & 0.062   & 0.045  & \textbf{0.033} \\
                           & S2                   & 22.64     & 23.61   & 25.80  & \textbf{26.09} & 0.852     & 0.864   & 0.891  & \textbf{0.914} & 0.165     & 0.165   & 0.135  & \textbf{0.135} \\
                           & S3                   & 21.46     & 22.60   & 24.99  & \textbf{25.85} & 0.779     & 0.797   & 0.835  & \textbf{0.881} & 0.226     & 0.207   & 0.169  & \textbf{0.164} \\
                           & S4                   & 20.53     & 21.61   & 24.15  & \textbf{25.65} & 0.686     & 0.706   & 0.758  & \textbf{0.816} & 0.295     & 0.265   & 0.207  & \textbf{0.197} \\ \hline
\multirow{4}{*}{Ficus}     & S1                   & 21.86     & 22.73   & 24.72  & \textbf{25.14} & 0.899     & 0.904   & 0.919  & \textbf{0.923} & 0.125     & 0.117   & 0.089  & \textbf{0.077} \\
                           & S2                   & 18.60     & 19.67   & 22.20  & \textbf{22.70} & 0.793     & 0.801   & 0.830  & \textbf{0.831} & 0.227     & 0.228   & 0.189  & \textbf{0.159} \\
                           & S3                   & 17.51     & 18.65   & 21.39  & \textbf{22.56} & 0.720     & 0.735   & 0.774  & \textbf{0.799} & 0.288     & 0.270   & 0.222  & \textbf{0.187} \\
                           & S4                   & 16.05     & 17.27   & 20.15  & \textbf{22.14} & 0.626     & 0.642   & 0.695  & \textbf{0.746} & 0.368     & 0.339   & 0.271  & \textbf{0.235} \\ \hline
\multirow{4}{*}{Materials} & S1                   & 19.47     & 19.93   & 20.98  & \textbf{22.67} & 0.828     & 0.847   & 0.895  & \textbf{0.903} & 0.195     & 0.174   & 0.124  & \textbf{0.110} \\
                           & S2                   & 14.41     & 14.91   & 16.48  & \textbf{19.15} & 0.654     & 0.683   & 0.739  & \textbf{0.807} & 0.357     & 0.348   & 0.288  & \textbf{0.198} \\
                           & S3                   & 13.15     & 13.92   & 15.66  & \textbf{18.54} & 0.581     & 0.615   & 0.683  & \textbf{0.774} & 0.420     & 0.390   & 0.322  & \textbf{0.227} \\
                           & S4                   & 11.89     & 12.66   & 14.58  & \textbf{16.25} & 0.486     & 0.521   & 0.603  & \textbf{0.729} & 0.514     & 0.472   & 0.386  & \textbf{0.267} \\ \hline
\multirow{4}{*}{Hotdog}    & S1                   & 22.14     & 24.70   & 30.45  & \textbf{32.70} & 0.902     & 0.919   & 0.958  & \textbf{0.968} & 0.135     & 0.115   & 0.066  & \textbf{0.054} \\
                           & S2                   & 17.20     & 20.02   & 26.29  & \textbf{28.82} & 0.797     & 0.817   & 0.871  & \textbf{0.904} & 0.275     & 0.259   & 0.198  & \textbf{0.149} \\
                           & S3                   & 16.22     & 18.99   & 25.48  & \textbf{28.20} & 0.724     & 0.750   & 0.814  & \textbf{0.871} & 0.332     & 0.301   & 0.232  & \textbf{0.178} \\
                           & S4                   & 14.98     & 17.79   & 24.43  & \textbf{26.27} & 0.636     & 0.662   & 0.740  & \textbf{0.825} & 0.410     & 0.365   & 0.276  & \textbf{0.216} \\ \hline
\multirow{4}{*}{Ave}       & S1                   & 21.20     & 22.47   & 25.31  & \textbf{26.45} & 0.857     & 0.874   & 0.913  & \textbf{0.922} & 0.161     & 0.143   & 0.104  & \textbf{0.092} \\
                           & S2                   & 17.00     & 18.44   & 21.80  & \textbf{23.16} & 0.732     & 0.755   & 0.805  & \textbf{0.842} & 0.295     & 0.286   & 0.236  & \textbf{0.183} \\
                           & S3                   & 15.88     & 17.43   & 20.99  & \textbf{22.70} & 0.660     & 0.687   & 0.749  & \textbf{0.810} & 0.355     & 0.328   & 0.270  & \textbf{0.211} \\
                           & S4                   & 14.67     & 16.25   & 19.97  & \textbf{21.72} & 0.567     & 0.597   & 0.672  & \textbf{0.758} & 0.440     & 0.400   & 0.322  & \textbf{0.248} \\ \hline
\end{tabular}}
\caption{Quantitative comparisons of methods (PixelNeRF \cite{yu2021pixelnerf}, MVSNeRF \cite{chen2021mvsnerf}, IBRNet \cite{wang2021ibrnet} and the proposed TransNeRF) for the scene-agnostic setting on the realistic synthetic dataset \cite{mildenhall2020nerf}. \label{Agn-realistic-synth}}
\end{table*}

\begin{table*}[]
\centering
\resizebox{\textwidth}{!}{
\begin{tabular}{cc|cccc|cccc|cccc} \hline
\multicolumn{14}{c}{Scene-agnostic setting on real forward-facing dataset \cite{mildenhall2019llff}}                                                                                                                       \\ \hline
\multicolumn{1}{l}{}      & \multicolumn{1}{l}{} & \multicolumn{4}{c}{PSNR $\uparrow$}              & \multicolumn{4}{c}{SSIM $\uparrow$}              & \multicolumn{4}{c}{LPIPS $\downarrow$}           \\ \hline
Scene                     & $\mathbf{S}_i$             & PixelNeRF & MVSNeRF & IBRNet & TransNeRF      & PixelNeRF & MVSNeRF & IBRNet & TransNeRF      & PixelNeRF & MVSNeRF & IBRNet & TransNeRF      \\ \hline
\multirow{3}{*}{Fern}     & S1                   & 20.65     & 21.12   & 23.69  & \textbf{23.70} & 0.671     & 0.696   & 0.767  & \textbf{0.771} & 0.355     & 0.322   & 0.250  & \textbf{0.247} \\
                          & S2                   & 18.69     & 19.40   & 22.18  & \textbf{22.35} & 0.602     & 0.639   & 0.709  & \textbf{0.720} & 0.427     & 0.389   & 0.307  & \textbf{0.295} \\
                          & S3                   & 17.99     & 18.80   & 21.96  & \textbf{22.11} & 0.574     & 0.607   & 0.703  & \textbf{0.704} & 0.460     & 0.410   & 0.316  & \textbf{0.304} \\ \hline
\multirow{3}{*}{Trex}     & S1                   & 18.63     & 19.24   & 23.83  & \textbf{23.84} & 0.705     & 0.722   & 0.849  & \textbf{0.850} & 0.392     & 0.377   & 0.239  & \textbf{0.237} \\
                          & S2                   & 16.03     & 16.88   & 21.68  & \textbf{21.85} & 0.633     & 0.660   & 0.788  & \textbf{0.797} & 0.467     & 0.448   & 0.299  & \textbf{0.287} \\
                          & S3                   & 13.80     & 15.18   & 19.94  & \textbf{20.51} & 0.561     & 0.611   & 0.737  & \textbf{0.764} & 0.544     & 0.516   & 0.352  & \textbf{0.328} \\ \hline
\multirow{3}{*}{Horns}    & S1                   & 20.01     & 21.35   & 26.34  & \textbf{26.35} & 0.702     & 0.739   & 0.866  & \textbf{0.868} & 0.356     & 0.323   & 0.179  & \textbf{0.179} \\
                          & S2                   & 16.78     & 18.37   & 23.56  & \textbf{23.74} & 0.626     & 0.680   & 0.802  & \textbf{0.812} & 0.438     & 0.401   & 0.247  & \textbf{0.236} \\
                          & S3                   & 14.06     & 15.99   & 21.32  & \textbf{21.71} & 0.537     & 0.610   & 0.734  & \textbf{0.760} & 0.531     & 0.482   & 0.316  & \textbf{0.293} \\ \hline
\multirow{3}{*}{Fortress} & S1                   & 22.37     & 24.31   & 29.97  & \textbf{29.98} & 0.719     & 0.732   & 0.879  & \textbf{0.880} & 0.327     & 0.287   & 0.155  & \textbf{0.152} \\
                          & S2                   & 20.14     & 22.36   & 28.19  & \textbf{28.40} & 0.683     & 0.701   & 0.853  & \textbf{0.862} & 0.368     & 0.325   & 0.180  & \textbf{0.170} \\
                          & S3                   & 15.15     & 17.97   & 23.68  & \textbf{24.36} & 0.560     & 0.595   & 0.751  & \textbf{0.767} & 0.498     & 0.446   & 0.287  & \textbf{0.267} \\ \hline
\multirow{3}{*}{Leaves}   & S1                   & 15.56     & 16.37   & 20.30  & \textbf{20.31} & 0.513     & 0.561   & 0.722  & \textbf{0.724} & 0.418     & 0.378   & 0.228  & \textbf{0.226} \\
                          & S2                   & 13.07     & 14.23   & 18.26  & \textbf{18.54} & 0.395     & 0.456   & 0.615  & \textbf{0.627} & 0.529     & 0.484   & 0.324  & \textbf{0.309} \\
                          & S3                   & 11.20     & 12.73   & 16.87  & \textbf{17.39} & 0.303     & 0.394   & 0.544  & \textbf{0.578} & 0.607     & 0.550   & 0.378  & \textbf{0.355} \\ \hline
\multirow{3}{*}{Orchids}  & S1                   & 15.57     & 16.14   & 19.25  & \textbf{19.26} & 0.464     & 0.492   & 0.629  & \textbf{0.631} & 0.461     & 0.422   & 0.294  & \textbf{0.291} \\
                          & S2                   & 13.64     & 14.58   & 17.76  & \textbf{18.07} & 0.371     & 0.410   & 0.547  & \textbf{0.558} & 0.560     & 0.517   & 0.378  & \textbf{0.363} \\
                          & S3                   & 12.28     & 13.42   & 16.89  & \textbf{17.26} & 0.299     & 0.345   & 0.496  & \textbf{0.507} & 0.627     & 0.575   & 0.421  & \textbf{0.399} \\ \hline
\multirow{3}{*}{Room}     & S1                   & 21.52     & 22.74   & 29.70  & \textbf{29.71} & 0.820     & 0.835   & 0.941  & \textbf{0.944} & 0.318     & 0.293   & 0.155  & \textbf{0.153} \\
                          & S2                   & 17.81     & 19.43   & 26.45  & \textbf{26.78} & 0.775     & 0.788   & 0.907  & \textbf{0.916} & 0.379     & 0.352   & 0.202  & \textbf{0.191} \\
                          & S3                   & 13.71     & 15.64   & 22.83  & \textbf{23.33} & 0.699     & 0.738   & 0.852  & \textbf{0.870} & 0.479     & 0.442   & 0.278  & \textbf{0.259} \\ \hline
\multirow{3}{*}{Flower}   & S1                   & 17.84     & 19.46   & 26.61  & \textbf{26.63} & 0.614     & 0.663   & 0.854  & \textbf{0.858} & 0.414     & 0.376   & 0.165  & \textbf{0.157} \\
                          & S2                   & 14.21     & 16.23   & 23.43  & \textbf{23.76} & 0.524     & 0.580   & 0.775  & \textbf{0.787} & 0.507     & 0.464   & 0.243  & \textbf{0.229} \\
                          & S3                   & 10.28     & 11.97   & 19.15  & \textbf{19.85} & 0.374     & 0.446   & 0.646  & \textbf{0.662} & 0.662     & 0.607   & 0.374  & \textbf{0.336} \\ \hline
\multirow{3}{*}{Ave}      & S1                   & 19.02     & 20.09   & 24.96  & \textbf{24.97} & 0.651     & 0.680   & 0.813  & \textbf{0.816} & 0.380     & 0.347   & 0.208  & \textbf{0.205} \\
                          & S2                   & 16.30     & 17.68   & 22.69  & \textbf{22.94} & 0.576     & 0.614   & 0.749  & \textbf{0.760} & 0.459     & 0.422   & 0.273  & \textbf{0.260} \\
                          & S3                   & 13.56     & 15.21   & 20.33  & \textbf{20.81} & 0.489     & 0.543   & 0.683  & \textbf{0.701} & 0.551     & 0.504   & 0.340  & \textbf{0.318} \\ \hline
\end{tabular}}
\caption{Quantitative comparisons of methods (PixelNeRF \cite{yu2021pixelnerf}, MVSNeRF \cite{chen2021mvsnerf}, IBRNet \cite{wang2021ibrnet} and the proposed TransNeRF) for the scene-agnostic setting on the real forward-facing dataset \cite{mildenhall2019llff}. \label{Agn-real}}
\end{table*}


\begin{table*}[]
\centering
\resizebox{\textwidth}{!}{
\begin{tabular}{cc|cccc|cccc|cccc} \hline
\multicolumn{14}{c}{Per-scene finetuning setting on realistic synthetic dataset \cite{mildenhall2020nerf}}                                                                                                                 \\ \hline
\multicolumn{1}{l}{}       & \multicolumn{1}{l}{} & \multicolumn{4}{c}{PSNR $\uparrow$}              & \multicolumn{4}{c}{SSIM $\uparrow$}              & \multicolumn{4}{c}{LPIPS $\downarrow$}           \\ \hline
Scene                      & $\mathbf{S}_i$             & PixelNeRF & MVSNeRF & IBRNet & TransNeRF      & PixelNeRF & MVSNeRF & IBRNet & TransNeRF      & PixelNeRF & MVSNeRF & IBRNet & TransNeRF      \\ \hline
\multirow{4}{*}{Chair}     & S1                   & 23.95     & 29.48   & 32.62  & \textbf{33.49} & 0.910     & 0.950   & 0.972  & \textbf{0.989} & 0.114     & 0.071   & 0.044  & \textbf{0.038} \\
                           & S2                   & 19.84     & 25.54   & 29.06  & \textbf{30.45} & 0.820     & 0.870   & 0.896  & \textbf{0.942} & 0.205     & 0.156   & 0.123  & \textbf{0.096} \\
                           & S3                   & 18.93     & 24.78   & 28.35  & \textbf{30.21} & 0.764     & 0.817   & 0.852  & \textbf{0.911} & 0.244     & 0.192   & 0.153  & \textbf{0.118} \\
                           & S4                   & 17.90     & 23.80   & 27.48  & \textbf{29.84} & 0.683     & 0.738   & 0.780  & \textbf{0.868} & 0.300     & 0.241   & 0.190  & \textbf{0.143} \\ \hline
\multirow{4}{*}{Lego}      & S1                   & 22.50     & 26.34   & 28.65  & \textbf{30.46} & 0.854     & 0.916   & 0.947  & \textbf{0.959} & 0.153     & 0.099   & 0.068  & \textbf{0.056} \\
                           & S2                   & 18.58     & 22.58   & 25.28  & \textbf{27.38} & 0.722     & 0.790   & 0.829  & \textbf{0.890} & 0.304     & 0.246   & 0.207  & \textbf{0.140} \\
                           & S3                   & 17.67     & 21.82   & 24.57  & \textbf{26.91} & 0.664     & 0.734   & 0.782  & \textbf{0.859} & 0.343     & 0.283   & 0.237  & \textbf{0.162} \\
                           & S4                   & 16.64     & 20.84   & 23.69  & \textbf{26.02} & 0.601     & 0.676   & 0.731  & \textbf{0.814} & 0.394     & 0.327   & 0.269  & \textbf{0.194} \\ \hline
\multirow{4}{*}{Ship}      & S1                   & 24.22     & 25.33   & 26.83  & \textbf{28.90} & 0.823     & 0.838   & 0.855  & \textbf{0.870} & 0.246     & 0.216   & 0.188  & \textbf{0.177} \\
                           & S2                   & 20.34     & 21.61   & 23.49  & \textbf{25.84} & 0.688     & 0.707   & 0.733  & \textbf{0.763} & 0.416     & 0.381   & 0.346  & \textbf{0.265} \\
                           & S3                   & 19.45     & 20.85   & 22.80  & \textbf{25.15} & 0.631     & 0.653   & 0.687  & \textbf{0.731} & 0.454     & 0.418   & 0.376  & \textbf{0.287} \\
                           & S4                   & 18.48     & 19.94   & 21.98  & \textbf{24.26} & 0.551     & 0.578   & 0.619  & \textbf{0.680} & 0.523     & 0.479   & 0.425  & \textbf{0.317} \\ \hline
\multirow{4}{*}{Drums}     & S1                   & 20.27     & 22.96   & 25.20  & \textbf{26.32} & 0.791     & 0.861   & 0.922  & \textbf{0.935} & 0.164     & 0.116   & 0.076  & \textbf{0.053} \\
                           & S2                   & 16.67     & 19.53   & 22.15  & \textbf{23.27} & 0.677     & 0.757   & 0.823  & \textbf{0.840} & 0.302     & 0.248   & 0.202  & \textbf{0.161} \\
                           & S3                   & 15.96     & 18.96   & 21.64  & \textbf{23.22} & 0.622     & 0.705   & 0.779  & \textbf{0.808} & 0.341     & 0.284   & 0.232  & \textbf{0.183} \\
                           & S4                   & 14.93     & 17.98   & 20.76  & \textbf{23.11} & 0.551     & 0.637   & 0.718  & \textbf{0.767} & 0.404     & 0.339   & 0.276  & \textbf{0.201} \\ \hline
\multirow{4}{*}{Mic}       & S1                   & 29.39     & 31.65   & 32.77  & \textbf{33.64} & 0.949     & 0.970   & 0.981  & \textbf{0.983} & 0.050     & 0.039   & 0.033  & \textbf{0.027} \\
                           & S2                   & 25.84     & 28.26   & 29.76  & \textbf{30.68} & 0.889     & 0.914   & 0.934  & \textbf{0.952} & 0.124     & 0.109   & 0.095  & \textbf{0.084} \\
                           & S3                   & 25.11     & 27.67   & 29.24  & \textbf{30.66} & 0.831     & 0.860   & 0.887  & \textbf{0.921} & 0.156     & 0.138   & 0.118  & \textbf{0.114} \\
                           & S4                   & 24.24     & 26.85   & 28.51  & \textbf{30.57} & 0.756     & 0.790   & 0.824  & \textbf{0.861} & 0.202     & 0.177   & 0.145  & \textbf{0.139} \\ \hline
\multirow{4}{*}{Ficus}     & S1                   & 24.74     & 26.68   & 28.54  & \textbf{29.46} & 0.919     & 0.933   & 0.949  & \textbf{0.957} & 0.104     & 0.076   & 0.051  & \textbf{0.044} \\
                           & S2                   & 21.80     & 23.90   & 26.15  & \textbf{27.29} & 0.825     & 0.844   & 0.869  & \textbf{0.872} & 0.203     & 0.171   & 0.137  & \textbf{0.119} \\
                           & S3                   & 21.07     & 23.32   & 25.62  & \textbf{27.19} & 0.771     & 0.792   & 0.826  & \textbf{0.838} & 0.242     & 0.206   & 0.167  & \textbf{0.141} \\
                           & S4                   & 19.81     & 22.11   & 24.51  & \textbf{27.06} & 0.695     & 0.721   & 0.762  & \textbf{0.791} & 0.295     & 0.253   & 0.202  & \textbf{0.184} \\ \hline
\multirow{4}{*}{Materials} & S1                   & 22.27     & 23.32   & 24.98  & \textbf{26.99} & 0.848     & 0.889   & 0.925  & \textbf{0.938} & 0.174     & 0.130   & 0.098  & \textbf{0.077} \\
                           & S2                   & 17.27     & 18.49   & 20.54  & \textbf{23.74} & 0.688     & 0.740   & 0.781  & \textbf{0.845} & 0.325     & 0.275   & 0.237  & \textbf{0.158} \\
                           & S3                   & 16.44     & 17.80   & 19.90  & \textbf{22.95} & 0.633     & 0.686   & 0.735  & \textbf{0.814} & 0.364     & 0.310   & 0.267  & \textbf{0.181} \\
                           & S4                   & 15.33     & 16.73   & 18.94  & \textbf{21.18} & 0.556     & 0.612   & 0.669  & \textbf{0.774} & 0.432     & 0.371   & 0.316  & \textbf{0.216} \\ \hline
\multirow{4}{*}{Hotdog}    & S1                   & 25.14     & 30.58   & 34.54  & \textbf{37.03} & 0.922     & 0.948   & 0.973  & \textbf{0.986} & 0.114     & 0.077   & 0.048  & \textbf{0.041} \\
                           & S2                   & 20.90     & 26.51   & 30.85  & \textbf{33.40} & 0.850     & 0.881   & 0.915  & \textbf{0.943} & 0.226     & 0.185   & 0.147  & \textbf{0.110} \\
                           & S3                   & 19.56     & 25.32   & 29.71  & \textbf{32.54} & 0.792     & 0.826   & 0.867  & \textbf{0.910} & 0.264     & 0.220   & 0.177  & \textbf{0.132} \\
                           & S4                   & 18.49     & 24.30   & 28.79  & \textbf{31.19} & 0.719     & 0.758   & 0.806  & \textbf{0.869} & 0.314     & 0.262   & 0.207  & \textbf{0.164} \\ \hline
\multirow{4}{*}{Ave}       & S1                   & 24.06     & 27.04   & 29.27  & \textbf{30.79} & 0.877     & 0.913   & 0.940  & \textbf{0.952} & 0.140     & 0.103   & 0.076  & \textbf{0.064} \\
                           & S2                   & 20.15     & 23.30   & 25.91  & \textbf{27.76} & 0.770     & 0.813   & 0.847  & \textbf{0.881} & 0.263     & 0.221   & 0.187  & \textbf{0.142} \\
                           & S3                   & 19.27     & 22.56   & 25.23  & \textbf{27.35} & 0.714     & 0.759   & 0.802  & \textbf{0.849} & 0.301     & 0.256   & 0.216  & \textbf{0.165} \\
                           & S4                   & 18.23     & 21.57   & 24.33  & \textbf{26.65} & 0.639     & 0.689   & 0.739  & \textbf{0.803} & 0.358     & 0.306   & 0.254  & \textbf{0.195} \\ \hline
\end{tabular}}
\caption{Quantitative comparisons of methods (PixelNeRF \cite{yu2021pixelnerf}, MVSNeRF \cite{chen2021mvsnerf}, IBRNet \cite{wang2021ibrnet} and the proposed TransNeRF) for the per-scene fine-tuning setting on the realistic synthetic dataset \cite{mildenhall2020nerf}. \label{Finetune-realistic-synth}}
\end{table*}


\begin{table*}[]
\centering
\resizebox{\textwidth}{!}{
\begin{tabular}{cc|cccc|cccc|cccc} \hline
\multicolumn{14}{c}{Per-scene fine-tuning setting on real forward-facing dataset \cite{mildenhall2019llff}}                                                                                                                \\ \hline
\multicolumn{1}{l}{}      & \multicolumn{1}{l}{} & \multicolumn{4}{c}{PSNR $\uparrow$}              & \multicolumn{4}{c}{SSIM $\uparrow$}              & \multicolumn{4}{c}{LPIPS $\downarrow$}           \\ \hline
Scene                     & $\mathbf{S}_i$             & PixelNeRF & MVSNeRF & IBRNet & TransNeRF      & PixelNeRF & MVSNeRF & IBRNet & TransNeRF      & PixelNeRF & MVSNeRF & IBRNet & TransNeRF      \\ \hline
\multirow{3}{*}{Fern}     & S1                   & 22.35     & 23.43   & 24.89  & \textbf{24.92} & 0.713     & 0.749   & 0.803  & \textbf{0.812} & 0.300     & 0.262   & 0.212  & \textbf{0.208} \\
                          & S2                   & 20.52     & 21.86   & 23.60  & \textbf{23.89} & 0.653     & 0.696   & 0.752  & \textbf{0.775} & 0.351     & 0.308   & 0.253  & \textbf{0.236} \\
                          & S3                   & 19.77     & 21.06   & 23.51  & \textbf{23.85} & 0.630     & 0.680   & 0.749  & \textbf{0.769} & 0.378     & 0.324   & 0.257  & \textbf{0.240} \\ \hline
\multirow{3}{*}{Trex}     & S1                   & 20.33     & 23.21   & 26.43  & \textbf{26.47} & 0.747     & 0.817   & 0.896  & \textbf{0.900} & 0.338     & 0.271   & 0.202  & \textbf{0.198} \\
                          & S2                   & 17.74     & 20.87   & 24.38  & \textbf{24.68} & 0.684     & 0.759   & 0.843  & \textbf{0.864} & 0.394     & 0.323   & 0.248  & \textbf{0.235} \\
                          & S3                   & 15.76     & 19.20   & 23.07  & \textbf{23.76} & 0.622     & 0.705   & 0.801  & \textbf{0.841} & 0.455     & 0.375   & 0.287  & \textbf{0.268} \\ \hline
\multirow{3}{*}{Horns}    & S1                   & 21.71     & 24.34   & 28.27  & \textbf{28.31} & 0.744     & 0.801   & 0.898  & \textbf{0.907} & 0.300     & 0.241   & 0.145  & \textbf{0.142} \\
                          & S2                   & 18.92     & 21.79   & 26.01  & \textbf{26.31} & 0.679     & 0.742   & 0.842  & \textbf{0.864} & 0.361     & 0.295   & 0.195  & \textbf{0.183} \\
                          & S3                   & 16.18     & 19.34   & 23.94  & \textbf{24.61} & 0.595     & 0.667   & 0.779  & \textbf{0.829} & 0.435     & 0.359   & 0.247  & \textbf{0.228} \\ \hline
\multirow{3}{*}{Fortress} & S1                   & 24.07     & 26.94   & 30.99  & \textbf{31.05} & 0.761     & 0.822   & 0.894  & \textbf{0.901} & 0.272     & 0.216   & 0.143  & \textbf{0.139} \\
                          & S2                   & 22.51     & 25.60   & 29.96  & \textbf{30.27} & 0.738     & 0.804   & 0.881  & \textbf{0.896} & 0.295     & 0.235   & 0.155  & \textbf{0.144} \\
                          & S3                   & 17.94     & 21.30   & 26.05  & \textbf{26.72} & 0.625     & 0.698   & 0.787  & \textbf{0.830} & 0.400     & 0.330   & 0.238  & \textbf{0.217} \\ \hline
\multirow{3}{*}{Leaves}   & S1                   & 17.26     & 19.22   & 21.65  & \textbf{21.69} & 0.555     & 0.637   & 0.773  & \textbf{0.776} & 0.363     & 0.283   & 0.189  & \textbf{0.186} \\
                          & S2                   & 14.78     & 16.97   & 19.70  & \textbf{19.99} & 0.432     & 0.523   & 0.659  & \textbf{0.691} & 0.450     & 0.364   & 0.267  & \textbf{0.256} \\
                          & S3                   & 12.80     & 15.33   & 18.39  & \textbf{19.10} & 0.351     & 0.447   & 0.598  & \textbf{0.634} & 0.512     & 0.415   & 0.305  & \textbf{0.284} \\ \hline
\multirow{3}{*}{Orchids}  & S1                   & 17.27     & 19.03   & 20.70  & \textbf{20.74} & 0.506     & 0.596   & 0.682  & \textbf{0.691} & 0.406     & 0.316   & 0.233  & \textbf{0.230} \\
                          & S2                   & 15.51     & 17.50   & 19.46  & \textbf{19.77} & 0.420     & 0.514   & 0.605  & \textbf{0.632} & 0.485     & 0.390   & 0.302  & \textbf{0.289} \\
                          & S3                   & 14.04     & 16.30   & 18.67  & \textbf{19.32} & 0.349     & 0.452   & 0.555  & \textbf{0.589} & 0.539     & 0.434   & 0.333  & \textbf{0.312} \\ \hline
\multirow{3}{*}{Room}     & S1                   & 23.22     & 27.59   & 32.03  & \textbf{32.08} & 0.862     & 0.904   & 0.955  & \textbf{0.960} & 0.263     & 0.205   & 0.140  & \textbf{0.137} \\
                          & S2                   & 19.95     & 24.56   & 29.30  & \textbf{29.59} & 0.828     & 0.875   & 0.930  & \textbf{0.947} & 0.304     & 0.243   & 0.172  & \textbf{0.160} \\
                          & S3                   & 16.56     & 21.42   & 26.58  & \textbf{27.20} & 0.766     & 0.820   & 0.888  & \textbf{0.920} & 0.378     & 0.306   & 0.222  & \textbf{0.203} \\ \hline
\multirow{3}{*}{Flower}   & S1                   & 19.54     & 22.77   & 27.89  & \textbf{27.93} & 0.656     & 0.740   & 0.872  & \textbf{0.878} & 0.359     & 0.284   & 0.148  & \textbf{0.145} \\
                          & S2                   & 16.27     & 19.74   & 25.15  & \textbf{25.45} & 0.567     & 0.657   & 0.792  & \textbf{0.818} & 0.429     & 0.348   & 0.209  & \textbf{0.196} \\
                          & S3                   & 12.24     & 15.00   & 20.79  & \textbf{21.46} & 0.414     & 0.512   & 0.659  & \textbf{0.721} & 0.565     & 0.473   & 0.322  & \textbf{0.299} \\ \hline
\multirow{3}{*}{Ave}      & S1                   & 20.72     & 23.32   & 26.61  & \textbf{26.65} & 0.693     & 0.758   & 0.847  & \textbf{0.853} & 0.325     & 0.260   & 0.177  & \textbf{0.173} \\
                          & S2                   & 18.28     & 21.11   & 24.69  & \textbf{24.99} & 0.625     & 0.696   & 0.788  & \textbf{0.811} & 0.384     & 0.313   & 0.225  & \textbf{0.212} \\
                          & S3                   & 15.66     & 18.62   & 22.62  & \textbf{23.25} & 0.544     & 0.623   & 0.727  & \textbf{0.767} & 0.458     & 0.377   & 0.276  & \textbf{0.256} \\ \hline
\end{tabular}}
\caption{Quantitative comparisons of methods (PixelNeRF \cite{yu2021pixelnerf}, MVSNeRF \cite{chen2021mvsnerf}, IBRNet \cite{wang2021ibrnet} and the proposed TransNeRF) for the per-scene finetuning setting on the real forward-facing dataset \cite{mildenhall2019llff}.
\label{Finetune-real}}
\end{table*}

\begin{figure*}[]
	\centering
	\includegraphics[width=1\textwidth]{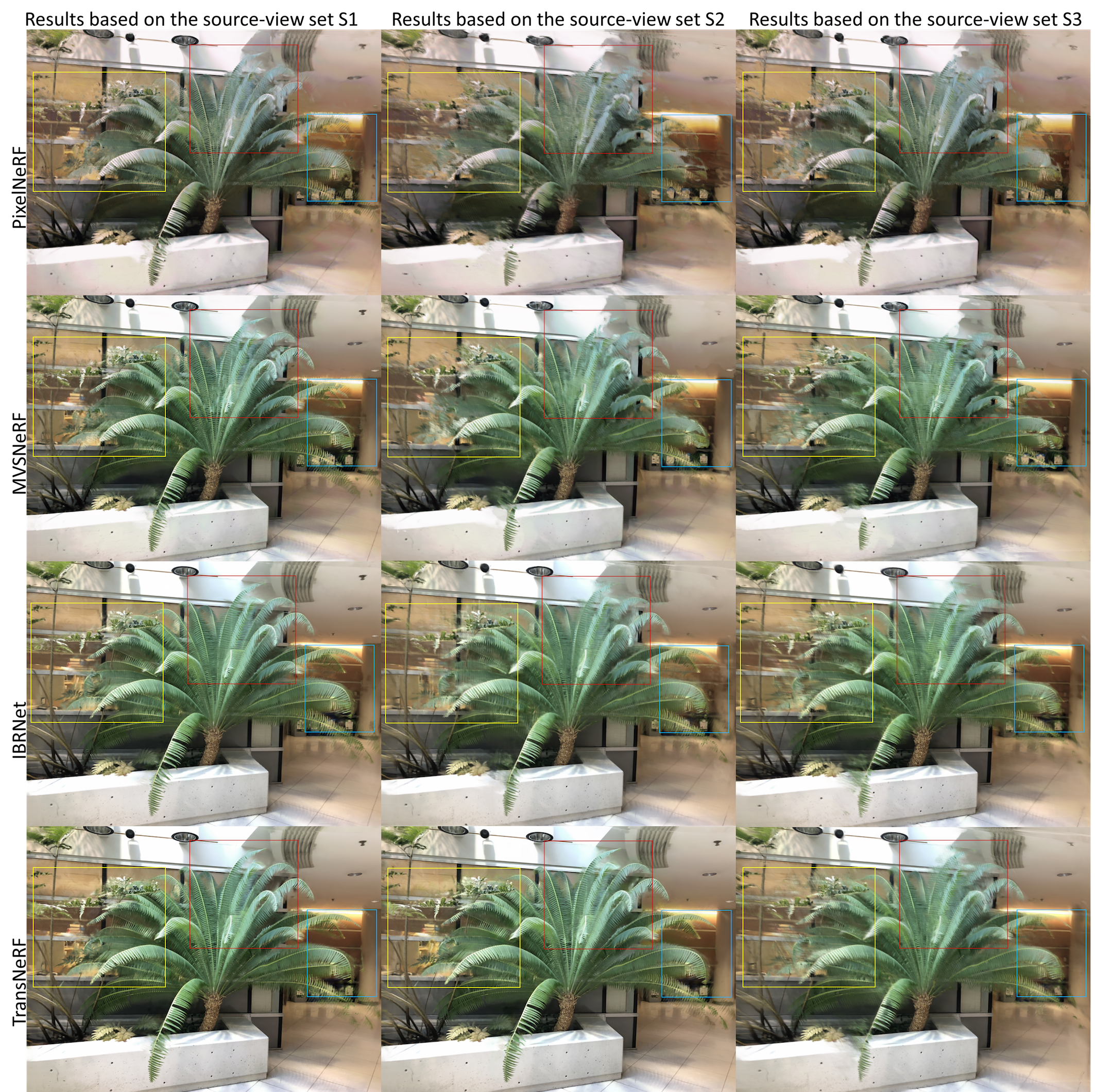}
	\caption{Qualitative results (PixelNeRF \cite{yu2021pixelnerf}, MVSNeRF \cite{chen2021mvsnerf}, IBRNet \cite{wang2021ibrnet} and our TransNeRF) for the fern scene \cite{mildenhall2019llff} under the scene-agnostic setting.}
	\label{fig:nonfine}
\end{figure*}
\begin{figure*}[]
	\centering
	\includegraphics[width=1\textwidth]{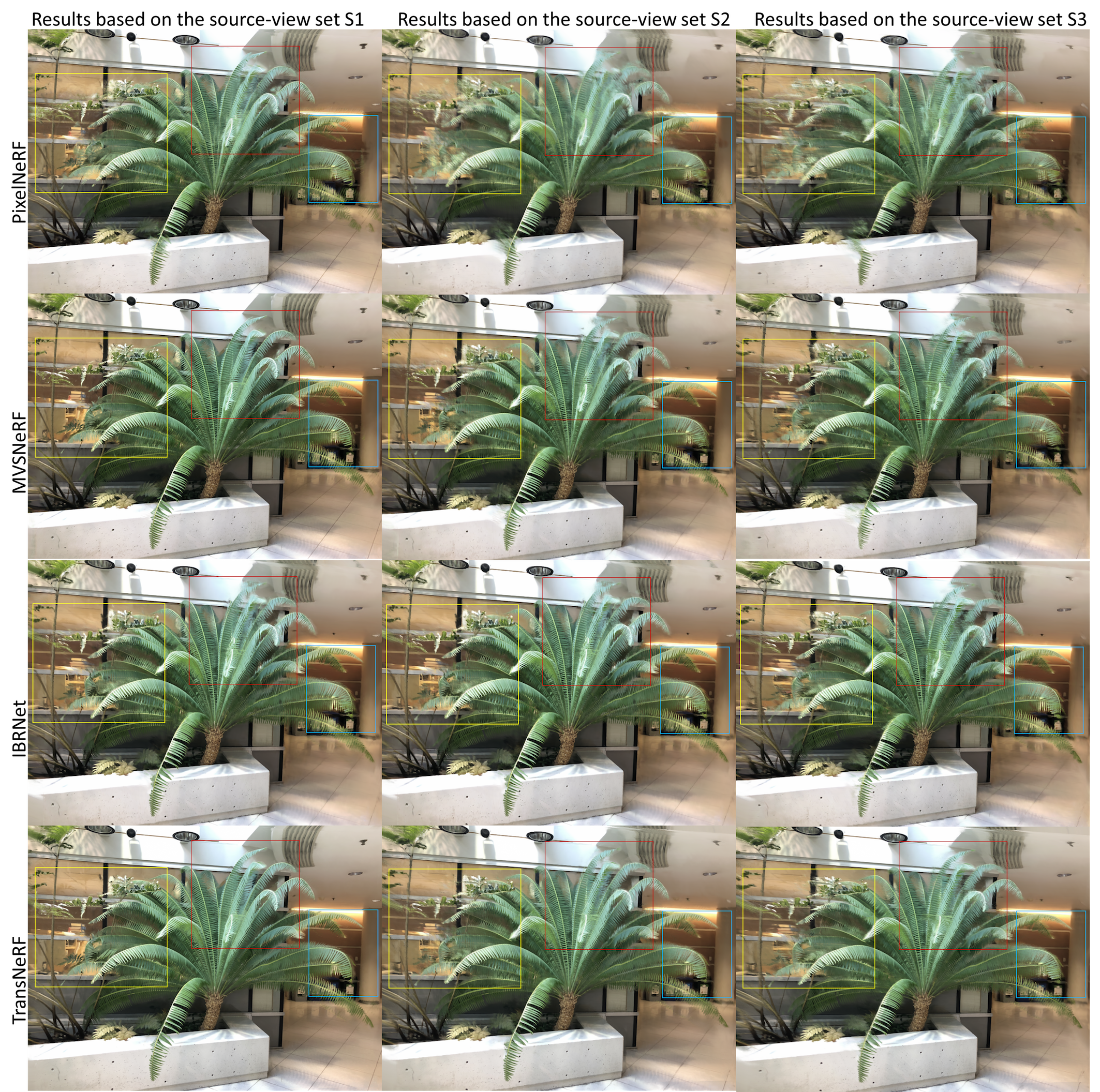}
	\caption{Qualitative results (PixelNeRF \cite{yu2021pixelnerf}, MVSNeRF \cite{chen2021mvsnerf}, IBRNet \cite{wang2021ibrnet} and our TransNeRF) for the fern scene \cite{mildenhall2019llff} under the per-scene finetuning setting.}
	\label{fig:fine}
\end{figure*}

\section{Experiments}
We evaluate our approach in the following experimental settings:
\begin{itemize}
\item Scene-agnostic setting: we train a single scene-agnostic model on a large training dataset, including various camera setups and scene types. We test its generalization ability to unseen-scene view synthesis on all test scenes from the evaluation dataset; 
\item Per-scene finetuning setting: our pre-trained scene-agnostic model can also be finetuned on each test scene. We then evaluate each scene-specific model on its corresponding scene separately.
\end{itemize}

The experiments are designed to examine whether our TransNeRF can efficiently learn a neural radiance field scene representation in scenarios where the difference between source views and the target rendering view varies.

\subsection{Datasets} \label{Datasets}
We train and evaluate our method on a collection of several multi-view datasets containing both synthetic data and real data, as in IBRNet \cite{wang2021ibrnet}.

\begin{itemize}
\item \textbf{Real-world datasets for training} include the Spaces dataset \cite{flynn2019deepview}, RealEstate10K \cite{zhou2018stereo}, and the handheld-cellphone-captured scene dataset. The Spaces dataset \cite{flynn2019deepview} has 100 scenes and each scene is collected with a 16 camera rig at 3 to 10 rig positions. RealEstate10K \cite{zhou2018stereo}, a large indoor-scene dataset, is captured from around 80K video clips with camera poses. The cellphone-captured scene dataset contains 95 real scenes (36 from LLFF \cite{mildenhall2019llff} and 59 from IBRNet \cite{wang2021ibrnet}), where each scene consists of 20 to 60 forward-facing images. 

COLMAP \cite{schonberger2016structure} is adopted to estimate camera poses, intrinsic parameters and scene bounds for real data as in \cite{mildenhall2020nerf}.

\item \textbf{Synthetic dataset for training} is generated by IBRNet \cite{wang2021ibrnet} from Google Scanned Objects, which contains 1,030 models with a variety of view density rates.
\end{itemize}

Similar to the sampling strategy in \cite{wang2021ibrnet}, we randomly sample $M$ source views for each target view from a pool of $m \times M$ views where $M$ is sampled uniformly at random from [8, 12] and $m$ is from [1, 5].

\begin{itemize}
\item \textbf{Real-world dataset for evaluation} collects 8 complex real-world scenes captured with a handheld cellphone (5 from LLFF \cite{mildenhall2019llff} and 3 from NeRF \cite{mildenhall2020nerf}). Each scene consists of 20 to 62 forward-facing images with $1008 \times 756$ pixels, and 1/8th of these is held out as the test set (7/8 for per-scene finetuning).
\item \textbf{Synthetic dataset for evaluation}, adopted from NeRF \cite{mildenhall2020nerf}, includes 8 objects with complicated geometry and realistic non-Lambertian materials which are rendered at $800 \times 800$ pixels from viewpoints sampled either on the upper hemisphere or full sphere (100 views for per-scene finetuning and 200 views for testing).
\end{itemize}

\textbf{Evaluation dependence on the difference between source views and the target view.}
For the evaluation of a specific scene, target rendering views are from the testing set of evaluation datasets, and the corresponding surrounding-view set is contained in the training set of evaluation datasets.
In this evaluation, we sampled M=10 views from the surrounding-view set as source views to render a target view.
And the sampling procedure of source views is defined as:
for each target rendering view, we first compute the differences between the target-view camera pose and the surrounding-view camera poses using Euclidean transformation matrices of camera poses \cite{mildenhall2020nerf,wang2021ibrnet}, and then rank the surrounding views according to how different they are, from small to large;
by this difference ranking, we construct $N_s$ source-view sets ($\lbrace\mathbf{S}_i\rbrace^{N_s}_{i=1}$) of 10 views from the surrounding-view set to render each test view. 
For the real-world evaluation dataset, there are $N_s=3$ sets, the top-10 views ($\mathbf{S}_1$), the middle-10 ($\mathbf{S}_2$), and the bottom-10 ($\mathbf{S}_3$) respectively; 
for the synthetic evaluation dataset, $N_s=4$, the top-10 ($\mathbf{S}_1$), the middle-10 ($\mathbf{S}_2$), the 3/4th-10 ($\mathbf{S}_3$), and the bottom-10 ($\mathbf{S}_4$).
This sampling strategy can be used to compare methods in scenarios where the camera poses of source views have diverse degrees
of difference, from small to large, relative to the camera pose of the target view.

\subsection{Metrics} For the task of novel view synthesis, we quantitatively evaluate the rendered image quality based on PSNR, SSIM \cite{wang2004image} (higher is better), and LPIPS \cite{zhang2018unreasonable} (lower is better) by comparing with the ground truth.

\subsection{Baselines}

We compare the proposed method quantitatively and qualitatively with top-performing generic NeRF approaches (PixelNeRF \cite{yu2021pixelnerf}, MVSNeRF \cite{chen2021mvsnerf}, IBRNet \cite{wang2021ibrnet}) that are also designed to generalize the scene-specific NeRF representation to a scene-agnostic one by taking multi-view features as conditional inputs.
In the evaluation, each comparing method trains a single set of model parameters on the same training dataset and tests their generalizability for novel scenes on the same evaluation dataset described in Sec. \ref{Datasets}, as do our model. 
Furthermore, for a specific scene, we also evaluate comparing methods in the per-scene finetuning experimental setting. 
All comparing approaches are implemented based on their released codes in the same training setting.

\subsection{Results}
In both the scene-agnostic and per-scene finetuning settings, we evaluate the performance of different methods in scenarios where the source views belong to the different source view sets defined in ($\lbrace\mathbf{S}_i\rbrace^{N_s}_{i=1}$) in Sec. \ref{Datasets}.
To render a testing view, each compared approach receives as input the same source-view set. 
We present both qualitative and quantitative results for evaluation. 
PSNR/SSIM (higher is better) and LPIPS (lower is better) are used as quantitative metrics. 
And for quantitative results, the best score for each category is in bold.
Tab. \ref{Agn-realistic-synth} and \ref{Agn-real} show quantitative results for the synthetic and real data in the scene-agnostic setting.
The quantitative results of the per-scene finetuning setting for the synthetic and real datasets are displayed in Tab. \ref{Finetune-realistic-synth} and \ref{Finetune-real}, respectively.
Qualitative results are shown in Fig. \ref{fig:nonfine} and \ref{fig:fine} for the scene-agnostic and per-scene finetuning settings, where the i-th column exhibits the rendering results for different methods based on the source view set $\mathbf{S}_i$ .

\subsubsection{Scene-agnostic Experiments}\label{Scene-agnostic Experiments}

In scene-agnostic experiments, we compare our TransNeRF with PixelNeRF \cite{yu2021pixelnerf}, MVSNeRF \cite{chen2021mvsnerf} and IBRNet \cite{wang2021ibrnet} on the real forward-facing dataset \cite{mildenhall2019llff} and the realistic synthetic dataset \cite{wang2021ibrnet}.

Tab. \ref{Agn-realistic-synth} and \ref{Agn-real} show that the proposed TransNeRF outperforms comparing methods in the scene-agnostic scenario on all test scenes from synthetic and real datasets.
It shows that our method has better generalization ability to novel scenes though it is trained on datasets with noticeably different scenes and view distributions.
The superior generalization ability of our method are also reflected in qualitative results. 
As shown in Fig. \ref{fig:nonfine}, we examine the performance of the comparing methods on rendering the same randomly-selected test view when receiving different source-view sets, and the ground-truth rendering view is displayed in Fig. \ref{fig:TransNeRF2}.
In Fig. \ref{fig:nonfine}, the results of comparing approaches contain more obvious artifacts than our approach and even completely fail in the $\mathbf{S}_3$ scenario, where the difference between the given source views and the target rendering view is more considerable than that in $\mathbf{S}_1$ and $\mathbf{S}_2$ scenarios.
As depicted in colored boxes, the comparing methods cannot synthesize clean boundary of fronds and recover thin structures (branches and leaves besides fern).
The artifacts become worse with the increase of the difference between the given source views and the target rendering view.

From the above qualitative results, we observe that there exists a gradual decline in the quality of the synthesized view when the difference between source views and the target rendering view increases from $\mathbf{S}_1$ to $\mathbf{S}_3$.
Similarly, in quantitative results from $\mathbf{S}_1$ to $\mathbf{S}_3$, PSNR and SSIM values both decrease while LPIPS increases for all comparing methods.
It reveals that the more dissimilar the source views are with respect to the target rendering view, the more difficult novel view synthesis becomes.
The tables also indicate that the advantage of TransNeRF becomes more significant than the comparing methods with the increase in the difference between source views and the target rendering view.
It demonstrates that TransNeRF has a stronger capability to explore complicated relationships between source views and the target view and learn a better scene representation in challenging scenarios.
Please see the supplemental material for more results.

\subsubsection{Per-scene Finetuning Experiments}
In the per-scene finetuning scenario, each pre-trained model of comparing methods (PixelNeRF \cite{yu2021pixelnerf}, MVSNeRF \cite{chen2021mvsnerf}, IBRNet \cite{wang2021ibrnet} and our TransNeRF) is finetuned for each scene in the evaluation dataset \cite{mildenhall2019llff, wang2021ibrnet}.

As shown in Tab. \ref{Finetune-realistic-synth} and \ref{Finetune-real}, TransNeRF outperforms other comparing methods after per-scene finetuning.
Similar to scene-agnostic results, per-scene finetuning results further validate that TransNeRF can learn more satisfactory novel view rendering than the comparing methods in the different source-view scenarios.
Meanwhile, the performance gap between TransNeRF and comparing methods becomes more extensive in contrast with that in the scene-agnostic setting, indicating that per-scene finetuning can further fulfill the potential of our TransNeRF.
Similar to quantitative results, Fig. \ref{fig:fine} illustrates that TransNeRF can achieve more realistic view synthesis results with fewer artifacts in comparison with baseline approaches.

\section{Conclusion} 
\label{sec:conclusion}

This paper proposes a unified Transformer-based framework to learn a general radiance field for novel view synthesis, which achieves state-of-the-art accuracy on both scene-agnostic and per-scene finetuning experimental settings for real and synthetic datasets. 
The proposed framework naturally utilizes the attention mechanism to integrate information of the projected 2D pixels in the surrounding source views and the neighboring 3D points along the query ray.
Meanwhile our framework explores and fuses deep and complicated relationships between observed views and the rendering view which is ignored in the previous MLP-based NeRFs.

{\small
 \bibliographystyle{IEEEtran}
 \bibliography{ref}

\begin{thebibliography}{10}
\providecommand{\url}[1]{#1}
\csname url@samestyle\endcsname
\providecommand{\newblock}{\relax}
\providecommand{\bibinfo}[2]{#2}
\providecommand{\BIBentrySTDinterwordspacing}{\spaceskip=0pt\relax}
\providecommand{\BIBentryALTinterwordstretchfactor}{4}
\providecommand{\BIBentryALTinterwordspacing}{\spaceskip=\fontdimen2\font plus
\BIBentryALTinterwordstretchfactor\fontdimen3\font minus
  \fontdimen4\font\relax}
\providecommand{\BIBforeignlanguage}[2]{{%
\expandafter\ifx\csname l@#1\endcsname\relax
\typeout{** WARNING: IEEEtran.bst: No hyphenation pattern has been}%
\typeout{** loaded for the language `#1'. Using the pattern for}%
\typeout{** the default language instead.}%
\else
\language=\csname l@#1\endcsname
\fi
#2}}
\providecommand{\BIBdecl}{\relax}
\BIBdecl

\bibitem{mildenhall2020nerf}
B.~Mildenhall, P.~P. Srinivasan, M.~Tancik, J.~T. Barron, R.~Ramamoorthi, and
  R.~Ng, ``Nerf: Representing scenes as neural radiance fields for view
  synthesis,'' in \emph{European conference on computer vision}.\hskip 1em plus
  0.5em minus 0.4em\relax Springer, 2020, pp. 405--421.

\bibitem{wang2021ibrnet}
Q.~Wang, Z.~Wang, K.~Genova, P.~P. Srinivasan, H.~Zhou, J.~T. Barron,
  R.~Martin-Brualla, N.~Snavely, and T.~Funkhouser, ``Ibrnet: Learning
  multi-view image-based rendering,'' in \emph{Proceedings of the IEEE/CVF
  Conference on Computer Vision and Pattern Recognition}, 2021, pp. 4690--4699.

\bibitem{8105827}
D.~M.~M. Rahaman and M.~Paul, ``Virtual view synthesis for free viewpoint video
  and multiview video compression using gaussian mixture modelling,''
  \emph{IEEE Transactions on Image Processing}, vol.~27, no.~3, pp. 1190--1201,
  2018.

\bibitem{6690212}
B.~Ham, D.~Min, C.~Oh, M.~N. Do, and K.~Sohn, ``Probability-based rendering for
  view synthesis,'' \emph{IEEE Transactions on Image Processing}, vol.~23,
  no.~2, pp. 870--884, 2014.

\bibitem{9382843}
N.~Meng, K.~Li, J.~Liu, and E.~Y. Lam, ``Light field view synthesis via
  aperture disparity and warping confidence map,'' \emph{IEEE Transactions on
  Image Processing}, vol.~30, pp. 3908--3921, 2021.

\bibitem{9222285}
A.~Palazzi, L.~Bergamini, S.~Calderara, and R.~Cucchiara, ``Warp and learn:
  Novel views generation for vehicles and other objects,'' \emph{IEEE
  Transactions on Pattern Analysis and Machine Intelligence}, vol.~44, no.~4,
  pp. 2216--2227, 2022.

\bibitem{7546862}
K.~Rematas, C.~H. Nguyen, T.~Ritschel, M.~Fritz, and T.~Tuytelaars, ``Novel
  views of objects from a single image,'' \emph{IEEE Transactions on Pattern
  Analysis and Machine Intelligence}, vol.~39, no.~8, pp. 1576--1590, 2017.

\bibitem{lingjie}
L.~Liu, M.~Habermann, V.~Rudnev, K.~Sarkar, J.~Gu, and C.~Theobalt, ``Neural
  actor: Neural free-view synthesis of human actors with pose control,''
  \emph{ACM Trans. Graph.}, vol.~40, no.~6, dec 2021.

\bibitem{gkioxari2019mesh}
G.~Gkioxari, J.~Malik, and J.~Johnson, ``Mesh r-cnn,'' in \emph{Proceedings of
  the IEEE/CVF International Conference on Computer Vision}, 2019, pp.
  9785--9795.

\bibitem{nash2020polygen}
C.~Nash, Y.~Ganin, S.~A. Eslami, and P.~Battaglia, ``Polygen: An autoregressive
  generative model of 3d meshes,'' in \emph{International Conference on Machine
  Learning}.\hskip 1em plus 0.5em minus 0.4em\relax PMLR, 2020, pp. 7220--7229.

\bibitem{qi2017pointnet}
C.~R. Qi, H.~Su, K.~Mo, and L.~J. Guibas, ``Pointnet: Deep learning on point
  sets for 3d classification and segmentation,'' in \emph{Proceedings of the
  IEEE conference on computer vision and pattern recognition}, 2017, pp.
  652--660.

\bibitem{yang2018dense}
B.~Yang, S.~Rosa, A.~Markham, N.~Trigoni, and H.~Wen, ``Dense 3d object
  reconstruction from a single depth view,'' \emph{IEEE transactions on pattern
  analysis and machine intelligence}, vol.~41, no.~12, pp. 2820--2834, 2018.

\bibitem{Wang_2021_ICCV}
D.~Wang, X.~Cui, X.~Chen, Z.~Zou, T.~Shi, S.~Salcudean, Z.~J. Wang, and
  R.~Ward, ``Multi-view 3d reconstruction with transformers,'' in
  \emph{Proceedings of the IEEE/CVF International Conference on Computer Vision
  (ICCV)}, October 2021, pp. 5722--5731.

\bibitem{mescheder2019occupancy}
L.~Mescheder, M.~Oechsle, M.~Niemeyer, S.~Nowozin, and A.~Geiger, ``Occupancy
  networks: Learning 3d reconstruction in function space,'' in
  \emph{Proceedings of the IEEE/CVF Conference on Computer Vision and Pattern
  Recognition}, 2019, pp. 4460--4470.

\bibitem{genova2020local}
K.~Genova, F.~Cole, A.~Sud, A.~Sarna, and T.~Funkhouser, ``Local deep implicit
  functions for 3d shape,'' in \emph{Proceedings of the IEEE/CVF Conference on
  Computer Vision and Pattern Recognition}, 2020, pp. 4857--4866.

\bibitem{park2019deepsdf}
J.~J. Park, P.~Florence, J.~Straub, R.~Newcombe, and S.~Lovegrove, ``Deepsdf:
  Learning continuous signed distance functions for shape representation,'' in
  \emph{Proceedings of the IEEE/CVF Conference on Computer Vision and Pattern
  Recognition}, 2019, pp. 165--174.

\bibitem{jiang2020local}
C.~Jiang, A.~Sud, A.~Makadia, J.~Huang, M.~Nie{\ss}ner, T.~Funkhouser
  \emph{et~al.}, ``Local implicit grid representations for 3d scenes,'' in
  \emph{Proceedings of the IEEE/CVF Conference on Computer Vision and Pattern
  Recognition}, 2020, pp. 6001--6010.

\bibitem{insafutdinov2018unsupervised}
E.~Insafutdinov and A.~Dosovitskiy, ``Unsupervised learning of shape and pose
  with differentiable point clouds,'' \emph{Advances in neural information
  processing systems}, vol.~31, 2018.

\bibitem{tulsiani2017multi}
S.~Tulsiani, T.~Zhou, A.~A. Efros, and J.~Malik, ``Multi-view supervision for
  single-view reconstruction via differentiable ray consistency,'' in
  \emph{Proceedings of the IEEE conference on computer vision and pattern
  recognition}, 2017, pp. 2626--2634.

\bibitem{sitzmann2020scene}
V.~Sitzmann, M.~Zollh{\"o}fer, and G.~Wetzstein, ``Scene representation
  networks: Continuous 3d-structure-aware neural scene representations,''
  \emph{Advances in Neural Information Processing Systems}, vol.~32, 2019.

\bibitem{niemeyer2020differentiable}
M.~Niemeyer, L.~Mescheder, M.~Oechsle, and A.~Geiger, ``Differentiable
  volumetric rendering: Learning implicit 3d representations without 3d
  supervision,'' in \emph{Proceedings of the IEEE/CVF Conference on Computer
  Vision and Pattern Recognition}, 2020, pp. 3504--3515.

\bibitem{yu2021pixelnerf}
A.~Yu, V.~Ye, M.~Tancik, and A.~Kanazawa, ``pixelnerf: Neural radiance fields
  from one or few images,'' in \emph{Proceedings of the IEEE/CVF Conference on
  Computer Vision and Pattern Recognition}, 2021, pp. 4578--4587.

\bibitem{chen2021mvsnerf}
A.~Chen, Z.~Xu, F.~Zhao, X.~Zhang, F.~Xiang, J.~Yu, and H.~Su, ``Mvsnerf: Fast
  generalizable radiance field reconstruction from multi-view stereo,'' in
  \emph{Proceedings of the IEEE/CVF International Conference on Computer
  Vision}, 2021, pp. 14\,124--14\,133.

\bibitem{trevithick2021grf}
A.~Trevithick and B.~Yang, ``Grf: Learning a general radiance field for 3d
  representation and rendering,'' in \emph{Proceedings of the IEEE/CVF
  International Conference on Computer Vision}, 2021, pp. 15\,182--15\,192.

\bibitem{martin2021nerf}
R.~Martin-Brualla, N.~Radwan, M.~S. Sajjadi, J.~T. Barron, A.~Dosovitskiy, and
  D.~Duckworth, ``Nerf in the wild: Neural radiance fields for unconstrained
  photo collections,'' in \emph{Proceedings of the IEEE/CVF Conference on
  Computer Vision and Pattern Recognition}, 2021, pp. 7210--7219.

\bibitem{goodfellow2014generative}
I.~Goodfellow, J.~Pouget-Abadie, M.~Mirza, B.~Xu, D.~Warde-Farley, S.~Ozair,
  A.~Courville, and Y.~Bengio, ``Generative adversarial nets,'' \emph{Advances
  in neural information processing systems}, vol.~27, 2014.

\bibitem{brown2020language}
T.~Brown, B.~Mann, N.~Ryder, M.~Subbiah, J.~D. Kaplan, P.~Dhariwal,
  A.~Neelakantan, P.~Shyam, G.~Sastry, A.~Askell \emph{et~al.}, ``Language
  models are few-shot learners,'' \emph{Advances in neural information
  processing systems}, vol.~33, pp. 1877--1901, 2020.

\bibitem{carion2020end}
N.~Carion, F.~Massa, G.~Synnaeve, N.~Usunier, A.~Kirillov, and S.~Zagoruyko,
  ``End-to-end object detection with transformers,'' in \emph{European
  Conference on Computer Vision}.\hskip 1em plus 0.5em minus 0.4em\relax
  Springer, 2020, pp. 213--229.

\bibitem{dosovitskiy2020image}
A.~Dosovitskiy, L.~Beyer, A.~Kolesnikov, D.~Weissenborn, X.~Zhai,
  T.~Unterthiner, M.~Dehghani, M.~Minderer, G.~Heigold, S.~Gelly, J.~Uszkoreit,
  and N.~Houlsby, ``An image is worth 16x16 words: Transformers for image
  recognition at scale,'' in \emph{International Conference on Learning
  Representations (ICLR)}, 2021.

\bibitem{chen2020generative}
M.~Chen, A.~Radford, R.~Child, J.~Wu, H.~Jun, D.~Luan, and I.~Sutskever,
  ``Generative pretraining from pixels,'' in \emph{International Conference on
  Machine Learning}.\hskip 1em plus 0.5em minus 0.4em\relax PMLR, 2020, pp.
  1691--1703.

\bibitem{9491931}
D.~Wang, X.~Cui, X.~Chen, R.~Ward, and Z.~J. Wang, ``Interpreting bottom-up
  decision-making of cnns via hierarchical inference,'' \emph{IEEE Transactions
  on Image Processing}, vol.~30, pp. 6701--6714, 2021.

\bibitem{kajiya1984ray}
J.~T. Kajiya and B.~P. Von~Herzen, ``Ray tracing volume densities,'' \emph{ACM
  SIGGRAPH computer graphics}, vol.~18, no.~3, pp. 165--174, 1984.

\bibitem{mildenhall2019llff}
B.~Mildenhall, P.~P. Srinivasan, R.~Ortiz-Cayon, N.~K. Kalantari,
  R.~Ramamoorthi, R.~Ng, and A.~Kar, ``Local light field fusion: Practical view
  synthesis with prescriptive sampling guidelines,'' \emph{ACM Transactions on
  Graphics (TOG)}, 2019.

\bibitem{flynn2019deepview}
J.~Flynn, M.~Broxton, P.~E. Debevec, M.~DuVall, G.~Fyffe, R.~S. Overbeck,
  N.~Snavely, and R.~Tucker, ``Deepview: View synthesis with learned gradient
  descent.'' in \emph{CVPR}, 2019.

\bibitem{zhou2018stereo}
T.~Zhou, R.~Tucker, J.~Flynn, G.~Fyffe, and N.~Snavely, ``Stereo magnification:
  learning view synthesis using multiplane images,'' \emph{ACM Transactions on
  Graphics (TOG)}, vol.~37, no.~4, pp. 1--12, 2018.

\bibitem{schonberger2016structure}
J.~L. Schonberger and J.-M. Frahm, ``Structure-from-motion revisited,'' in
  \emph{Proceedings of the IEEE conference on computer vision and pattern
  recognition}, 2016, pp. 4104--4113.

\bibitem{wang2004image}
Z.~Wang, A.~C. Bovik, H.~R. Sheikh, and E.~P. Simoncelli, ``Image quality
  assessment: from error visibility to structural similarity,'' \emph{IEEE
  transactions on image processing}, vol.~13, no.~4, pp. 600--612, 2004.

\bibitem{zhang2018unreasonable}
R.~Zhang, P.~Isola, A.~A. Efros, E.~Shechtman, and O.~Wang, ``The unreasonable
  effectiveness of deep features as a perceptual metric,'' in \emph{Proceedings
  of the IEEE conference on computer vision and pattern recognition}, 2018, pp.
  586--595.

\end{thebibliography}
}

\end{document}